\documentclass{article}
\usepackage{ijcai26}

\usepackage{times}
\usepackage{soul}
\usepackage{url}
\usepackage[hidelinks]{hyperref}
\usepackage[utf8]{inputenc}
\usepackage[small]{caption}
\usepackage{graphicx}
\usepackage{amsmath}
\usepackage{amsthm}
\usepackage{booktabs}
\usepackage{algorithm}
\usepackage[switch]{lineno}
\usepackage{natbib}

\usepackage{amsfonts}       
\usepackage{nicefrac}       
\usepackage{xcolor}         
\usepackage{mathtools}
\usepackage{graphicx}
\usepackage{amssymb}
\usepackage{bm}
\usepackage{algorithm}
\usepackage[noend]{algpseudocode}
\usepackage{subcaption}
\usepackage[inkscapelatex=false]{svg}
\usepackage{multirow}  
\usepackage{tikz}
\usetikzlibrary{arrows.meta,positioning}

\usepackage[toc,page]{appendix}
\usepackage{minitoc}

\usepackage{longtable}

\newtheorem{theorem}{Theorem}

\newtheorem{definition}{Definition}
\newcommand{\firl}{$f$-IRL}
\newcommand{\tirl}{\textsf{TraIRL}}

\title{Inversely Learning Transferable Rewards via Abstracted States}

\author{
Yikang Gui$^1$
\And
Prashant Doshi$^2$\\
\affiliations
$^1$THINC Lab, School of Computing\\
University of Georgia, 
Athens, GA 30602\\
$^2$ School of Computing and Institute for AI\\
University of Georgia, 
Athens, GA 30602\\
\emails
\{yikang.gui, pdoshi\}@uga.edu
}

\begin{document}

\maketitle

\begin{abstract}
    Inverse reinforcement learning (IRL) has made significant progress in inferring reward functions from expert demonstrations. However, a key challenge to its use remains: how can we learn reward functions that generalize across related but distinct tasks. In this paper, we address this by focusing on {\em transferable} IRL, learning intrinsic rewards that can drive effective behavior in unseen but structurally aligned environments. Our method relies on a variational autoencoder to learn an abstract representation of the state space shared across multiple source tasks. This abstract space captures high-level features that are invariant between tasks, allowing the learning of a unified abstract reward function. The learned reward is then used to train policies in a separate, previously unseen target task without requiring new demonstrations in the target task. We evaluate our approach in multiple environments from Gymnasium and AssistiveGym, demonstrating that the learned abstract rewards consistently support successful policy learning in novel task settings.
\end{abstract}

\doparttoc 
\faketableofcontents 

\section{Introduction}
\label{sec:intro}

    The objective of inverse reinforcement learning (IRL) is one of abductive reasoning: to infer the reward function that best explains observed expert trajectories. This is challenging because available data are often sparse, admitting many potential solutions (some degenerate), and the learned reward functions often fail to generalize when transferred to other tasks with distribution shifts. Despite these challenges, significant progress has been made in the last decade toward learning the underlying reward functions in discrete and continuous domains~\citep{Arora21:Survey}. A central challenge in IRL is to learn reward functions that capture {\em task-invariant behavioral structure}: preferences that remain relevant across aligned tasks unseen during training. This contributes to the transferability of learned rewards, a characteristic of a general solution and key to IRL's appeal over related techniques such as behavior cloning.

    In this paper, we introduce a new method that generalizes IRL to unseen tasks that exhibit commonality with the observed ones in terms of shared intrinsic preferences. Abstractions offer a powerful representation toward generalization \citep{state-abstraction-rl}, and we introduce the novel concept of an {\em abstract reward function}. To illustrate, consider the Ant domain from OpenAI Gymnasium~\citep{schulman2018highdimensional}. Let two Ant environments with differing pairs of disabled legs act as source environments and an Ant environment with another pair of disabled legs as the target. As the source and target ants have different disabled legs, the marginal state distributions of the sources differ from the target's, which makes it difficult to transfer a learned reward function. However, if we focus on the ant's torso instead of its legs, the marginal state distribution of the torso remains largely invariant across both the sources and the target environment. Consequently, recovering a reward function based on the torso, which is the \textit{abstraction}, allows the function to be transferred across varying morphologies.

    Our method leverages observed behavior data from two or more tasks as input to a specialized variational autoencoder (VAE) framework. A single encoder is coupled with multiple task-specific decoders to reconstruct the trajectories unique to each source instance. We demonstrate how the common latent variable(s) of this VAE model can be interpreted and shaped as an abstract reward function that governs the task behaviors across instances of the domain. Crucially, two or more aligned task behaviors are required to disentangle shared task-invariant behavioral structure from task-specific dynamics. Furthermore, we incorporate a discriminator to regularize the latent space, ensuring it is structured by optimality information (e.g., distinguishing expert from suboptimal trajectories). This adversarial component is important because it forces the latent representation to prioritize features that are salient to the reward signal, rather than in mere reconstruction of state-action pairs.

    We evaluate our method for {\em transferable IRL}, labeled \tirl{}, on two benchmarks: Gymnasium domains \citep{schulman2018highdimensional} and AssistiveGym for collaborative robots \citep{DBLP:journals/corr/abs-1910-04700}. We utilize trajectories from multiple tasks in each domain as input to the VAE and show how the inversely learned abstract reward function can help successfully learn a high-quality behavior in another aligned task of the domain or even for a different domain. These results open a new frontier for methods that can learn abstract rewards via IRL to offer a level of generalizability not previously seen in the literature. 

\section{Related Work}
\label{sec:related}

    \textbf{Extant transfer learning for IRL struggles with mismatches in environment dynamics, which limits reward transferability.}
    \citet{transfer-irl} introduce an approach to learn diverse strategies from multiple experts by exploiting shared knowledge. However, it assumes unchanged dynamics between experts, which limits its applicability to settings with dynamics mismatch.
    I2L~\citep{Gangwani020} is designed for state-only imitation learning and addresses transition dynamics mismatches by using a prioritized trajectory buffer and by optimizing a lower bound on the expert's state-action visitation distribution. While empirically effective, it lacks theoretical guarantees on reward transferability and does not formally justify how the learned reward generalizes across different dynamics.
    \citet{10.5555/3540261.3542245} analyzes maximum causal entropy (MCE) IRL under transition dynamics mismatch, deriving necessary and sufficient conditions for transferability and providing a tight bound on performance degradation. It offers a robust MCE-based IRL algorithm but struggles to generalize under action-space shifts due to its reliance on matching state-action occupancy measures.
    \citet{yoo2022} studies transferable reward learning from multi-task demonstrations. However, their framework learns a latent subtask-conditioned reward and policy through {\em a posteriori} subtask variable whereas \tirl{} learns a shared cross-task state abstraction and a single stationary reward over the learned abstract manifold.

    \textbf{Rewards learned by adversarial IRL and IL methods may not transfer across environments}. AIRL~\citep{airl}, $f$-MAX~\citep{f-max} and $f$-IRL~\citep{firl} claim that their learned reward functions generalize to unseen or dynamically different environments. But these claims are not supported by explicit structural frameworks or theoretical guarantees, which leaves the transferability unpredictable. Furthermore, the learned rewards are tied to specific expert policy trajectories, which prevents their use in training new policies from scratch. In contrast, IQ-Learn \citep{iq-learn} is non-adversarial and learns soft Q-functions from expert data, which offers improved stability and efficiency. However, its reliance on Q-functions, which are action dependent, limits generalization to state-only reward functions. Unlike meta-IRL approaches which condition on latent task variables, \tirl{} learns a single stationary reward function intended for deployment without task inference or adaptation.

    \textbf{Reward identifiability remains a central theoretical challenge in IRL.}
    A fundamental ambiguity in IRL is that multiple reward functions can induce the same expert behavior, making the expert's underlying reward generally non-identifiable without additional assumptions. Prior theoretical work studies this issue from complementary perspectives. \citet{kim2021} analyze reward identifiability in deterministic MDPs under the MaxEntRL objective, \citet{rolland2022identifiability} characterize when rewards become identifiable or generalizable from multiple experts under structured MDP assumptions, and \citet{schlaginhaufen2024towards} provide conditions under which rewards recovered by regularized IRL remain transferable under changes in transition laws. \tirl{} is orthogonal to these identifiability-focused analyses: rather than claiming full recovery of the ground-truth reward, it targets a transferable abstract reward. Specifically, \tirl{} uses multi-task demonstrations to learn a shared cross-task state abstraction and then recovers a single stationary state-only reward over the learned abstract manifold. 

    \textbf{Existing successor feature matching algorithms lack transferable feature functions, which limits their ability to generalize to unseen tasks.} SFM~\citep{sfm} introduces successor feature matching into IRL while avoiding adversarial training. However, its feature functions are not designed for transfer learning. Our method can be viewed as an extension of SFM, where we incorporate a transferable abstract feature function. In Appendix~\ref{appendix:other_irl} in the supplement, we compare SFM as a backbone with \firl{} as a backbone.
    Crucially, \tirl{} differs from SFM-based approaches in that the abstraction itself is learned jointly from multiple tasks and is explicitly optimized for reward transfer, rather than being treated as a fixed feature space.

\section{Background}
\label{sec:background}

We briefly review MCE based IRL~\citep{mce-irl} as it informs our method. The entropy-regularized Markov decision process (MDP) is characterized by the tuple $(\mathcal{S}, \mathcal{A}, \mathcal{T}, r, \gamma, \rho_0)$. Here, $\mathcal{S}$ and $\mathcal{A}$ denote the state and action spaces, respectively, while $\gamma\in(0, 1)$ is the discount factor. In the standard IRL context, the dynamics modeled by the transition distribution $\mathcal{T}(s^\prime | a, s)$, the initial state distribution $\rho_0(s)$, and the reward function $r(s, a)$ are not known, and can only be determined through interaction with the environment. Optimal policy $\pi$ under the maximum entropy framework maximizes the objective 
$       \pi^* = \arg\max_\pi \mathbb{E}_{\tau\sim\pi}\left[\sum_{t=0}^T \gamma^t(r(s_t,a_t)+ H(\pi(\cdot|s_t)))\right]$, where $\tau \triangleq (s_0,a_0,s_1,a_1,\ldots,s_T,a_T)$ denotes a sequence of states and actions induced by the policy and transition function, and $H(\pi(\cdot|s))$ is the entropy of the action distribution due to policy $\pi$ for state $s$.
    
 The method \firl{}~\citep{firl} integrates $f$-divergence to improve scalability and robustness. \firl{} relies on a generator-discriminator schema to recover a stationary reward function by matching the expert's state marginal distribution (also called {\em state density} or occupancy distribution). This approach builds on and improves on the state marginal matching (SMM) algorithm~\citep{smm} for IRL. 
 We rely on a variant of $f$-IRL that minimizes the 1-Wasserstein distance between the state marginals, as this distance is an integral probability metric:
    \begin{equation}
        \mathcal{L}_\mathcal{F}({\bm \theta}) = \mathcal{D}_\mathcal{F}(\rho_E || \rho_{\bm \theta}) = W_1(\rho_E(s)\, , \,\rho_{\bm \theta}(s)),
        \label{eqn:firl-ipm-objective}
    \end{equation}
where $\mathcal{D}_\mathcal{F}$ is a divergence measure between distributions, $W_1$ is the 1-Wasserstein distance, $\rho_E$ and $\rho_{\bm \theta}$ denote the state densities of the expert and the soft-optimal learner under the reward function $R_{\bm \theta}$. Another advantage of \firl{} is its separation of the reward and discriminator networks, effectively forming a distillation model. The discriminator serves as the stronger model, while the reward function distills its information yielding better generalization than the single-discriminator design used in most adversarial IRL methods.

\section{Learning Transferable Rewards via Abstraction}
\label{sec:TIRL}

\begin{figure*}[!t]
      \centering
       \includegraphics[width=0.95\textwidth]{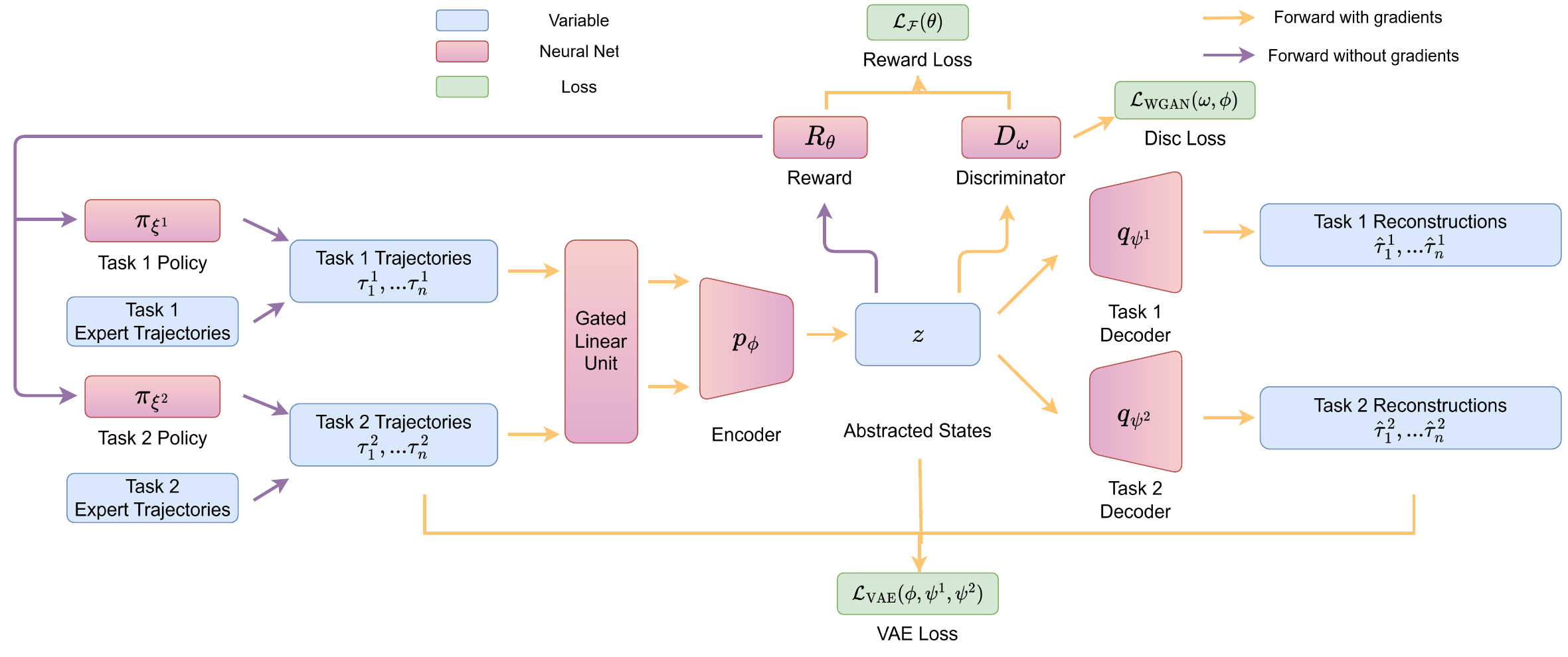}
       \caption{Overview of \tirl{}. Expert and learner trajectories in multiple source tasks are mapped to shared abstract states via a shared encoder. A discriminator compares the abstracted state densities to estimate the 1-Wasserstein distance between the expert and learner, which guides learning a reward function over the abstract space through a covariance-based objective. The learned reward function then optimizes the learner policy.}
       \label{fig:tirl-diagram}
    \end{figure*}

We present \tirl{} in this section and its schematic in Figure~\ref{fig:tirl-diagram}. The approach learns a state-only abstracted reward function optimized for general transfer from expert trajectories in source tasks to unseen target environments. This abstract reward function is leveraged to train a policy in the target domain using RL without additional expert trajectories.

\subsection{Problem Definition}

   We are provided with a set of expert trajectories which may be induced from policies of multiple source MDPs \(\{ \mathcal{M}^i \}_{i=1}^n\), each corresponding to a distinct but related task. The goal is to infer a shared intrinsic reward function that generalizes to a previously unseen target MDP \(\mathcal{M}_T\). Critically, this allows an agent to perform the task effectively without access to expert demonstrations in the target task.
    
    To support such transfer, we define a shared \emph{abstracted state space} \(\bar{\mathcal{S}}\) that captures task-invariant features across the source MDPs. This is formalized below:
    
    \begin{definition}[Cross-task abstraction] \label{def:cross_task_abstraction}
    The ground MDP for task \(i\) is defined as \(\mathcal{M}^i = (\mathcal{S}^i, \mathcal{A}^i, \mathcal{P}^i, \mathcal{R}^i, \gamma^i, \rho^i_0)\).
    A cross-task abstraction is defined by a mapping \(\phi: \mathcal{S}^i \rightarrow \bar{\mathcal{S}}\), where \(\phi(s^i) \in \bar{\mathcal{S}}\) denotes the abstracted state corresponding to the ground state \(s^i\). The inverse mapping \(\psi^i(\bar{s})\) denotes the set of ground states of task \(i\) that are mapped to the abstract state \(\bar{s} \in \bar{\mathcal{S}}\). 
    \end{definition}

    \tirl{} assumes that the source and target tasks share a common abstract manifold though this manifold is unknown {\em a priori} and learned from source demonstrations. Our objective is to learn an abstract reward function \(\bar{\mathcal{R}}: \bar{\mathcal{S}} \rightarrow \mathbb{R}\) and a cross-task abstraction \(\phi\) such that the composition $\bar{\mathcal{R}}\circ\phi:\mathcal{S}^i\rightarrow\mathbb{R}$ induces expert behaviors across source tasks $i=1$ to $n$ when used to substitute $\mathcal{R}^i$ in $\mathcal{M}^i$. The abstract reward function $\bar{\mathcal{R}}$ and the cross-task abstraction \(\phi\) are then utilized to guide policy search in the target MDP $\mathcal{M}^T$, assuming that $\mathcal{M}^T$ shares the same abstract manifold $\bar{\mathcal{S}}$. 

\subsection{Learning Abstraction via Multi-Head VAE}
    \label{subsec:VAE}

    To enable reward transfer across  tasks, we extract an abstract representation that captures the intrinsic, task-invariant structure from expert demonstrations. In \tirl{}, we realize the cross-task abstraction function $\phi$ using the encoder of a VAE. Specifically, we employ a \emph{single} task-agnostic encoder \( p_\phi\) shared across all source tasks and \emph{n} task-specific decoders \(\{q_{\bm \psi^i}\}_{i=1}^n\). The encoder \(p_\phi(z|s)\) maps a ground state \(s\in\mathcal{S}^i\) to a latent variable $z\in \bar{\mathcal{S}}$, which serves as the abstracted state $\phi(s)$.  The corresponding decoder \(q_{\bm \psi}(s|z)\) approximates the inverse mapping \(\psi(z)\), reconstructing the original ground state from its abstracted state.

    To facilitate the distillation of task-invariant features, we incorporate a gated linear unit (GLU) at the input layer of the encoder. For any given ground state \(s\), the gated transformation is defined as:
    \begin{align}
        \tilde{s} = \sigma(s V + b_v) \odot (s W + b_w)
    \end{align}
    While the gating mechanism enables the encoder to dynamically mask task-specific noise, the multi-task reconstruction objective is critical for maintaining the topological continuity of the abstracted state manifold. Without this reconstruction constraint, the manifold risks collapsing into disjoint clusters as we discuss in Section~\ref{subsec:density_ratio}. Such collapse would result in excessive separation between expert trajectories within the abstracted space, creating a discontinuity that hinders the learner policy from effectively sampling and reaching high-reward states. By enforcing ground state reconstruction, the encoder preserves a dense and continuous abstracted state manifold that bridges the gap between agent exploration and expert behavior.
    
    For our multi-task setting, the overall VAE objective is
    \begin{small}
    \begin{align}
         &\mathcal{L}_{\text{VAE}}({\bm \phi}, \bm \psi^1, \ldots, \bm \psi^n; {\bm \tau}) = \sum\nolimits_{i=1}^n \mathcal{L}_{\text{VAE}}^i({\bm \phi}, \bm \psi^i; {\bm \tau^i}) = \nonumber\\
         & \sum_{i=1}^n ~\mathbb{E}_{z \sim p_{\bm \phi}(z|s^{i})}~\log q_{\psi^{i}}(s^{i}|z) 
         - \lambda_{\mathcal{D}}\mathcal{D}_{\text{KL}} \left( p_{\bm \phi}(z|s^{i})\| p(z) \right)
        \label{eqn:vae_objective}
    \end{align}
    \end{small}
    where \({\bm \tau} \triangleq \langle \tau^1, \ldots, \tau^n \rangle\) denotes the collection of expert trajectories from \(n\) source tasks and each trajectory \(\tau^i = \{s^i_0, s^i_1, \ldots, s^i_T\}\) consists of a sequence of states from the \(i\)-th source environment. In the rest of the paper, $s_t^i$ denotes the state at time step $t$ from task $i$, and $\tau^i$ denotes a trajectory from task $i$. The prior \(p(z)\) is a normal distribution \(\mathcal{N}(0, I)\), and \(\lambda_{\mathcal{D}}\) controls the weight of the KL regularization term.

    \subsection{Structuring the Abstract State Space}
    \label{subsec:density_ratio}

    While the VAE objective ensures a representative abstraction through reconstruction, it is optimality-agnostic as it preserves environmental features without distinguishing expert behaviors from suboptimal trajectories. To ensure that the abstracted state space is semantically structured for reward recovery, we introduce a discriminator-guided regularization mechanism. This allows the manifold to become optimality-aware by explicitly embedding the distributional differences between expert and learner occupancy measures.

    We employ Wasserstein GAN with spectral normalization to estimate the 1-Wasserstein distance between the abstracted state distributions of the expert and the learner. The objective function of WGAN in our multi-task setting is
    \begin{small}
    \begin{align} \label{eqn:wgan_objective}
        & \mathcal{L}_{\text{WGAN}}({\bm \phi, \bm \omega \, ; \bm \tau}) = \sum\nolimits_{i=1}^n\mathcal{L}_{\text{WGAN}}^i({\bm \phi, \bm \omega \, ; \bm \tau^i}) = \nonumber\\ 
        & \sum\nolimits_{i=1}^n ~\mathbb{E}_{p_{\bm \phi}(z|s), \rho_L(s^i)} [D_{\bm \omega}(z)]
        - \mathbb{E}_{p_{\bm \phi}(z|s), \rho_E(s^i)}[D_{\bm \omega}(z)]
    \end{align}
    \end{small}
    where $\rho_L(s^i)$ and $\rho_E(s^i)$ denote the state densities of the learner and expert in the $i$-th source task, respectively, and \(D_{\bm \omega}\) denotes the discriminator. By distinguishing between these trajectories, the discriminator imposes an optimality-relevant topology on the abstracted state space, which facilitates the learning of a generalizable reward function.

    \subsection{Robust Transferable Reward Learning via Abstracted States}
    \label{subsec:gradient}
        Once the abstracted state manifold is structured, we recover a shared reward function \(R_{\bm \theta}(z)\) defined over this manifold. It is important to note that \tirl{} is not strictly coupled to a specific IRL framework; however, we adopt \firl{} as our underlying backbone. This choice is motivated by the functional decoupling of the reward and discriminator networks, which provides two primary advantages for transfer learning.

        First, the separation allows the reward function to act as a stationary optimization target, analogous to a target network in deep reinforcement learning. By decoupling the reward from the high-variance fluctuations of the adversarial discriminator, we ensure the recovered reward is robust to data variance and provides a stable signal for the learner policy. Second, this architecture induces a natural distillation process. By forcing the reward function to approximate the density misalignments captured by the discriminator, we filter out task-specific artifacts and overfitted features, thus ensuring that only generalizable, task-invariant intent is preserved.
        
        In our multi-task adaptation, the reward function \(R_{\bm \theta}(z)\) is trained to minimize the 1-Wasserstein distance between the abstracted state densities of the expert and the learner. The objective function is defined as:
        \begin{align} \label{eqn:trairl-reward-objective}
            & \mathcal{L}_\mathcal{F}({\bm \theta}) = \sum\limits_{i=1}^n W_1(\rho_E(z^i) , \rho_L(z^i)) = \nonumber\\
            & \sum\limits_{i=1}^n ~\mathbb{E}_{p_{\bm \phi}(z|s), \rho_E(s^i)} [D_{\bm \omega}(z)] 
         - \mathbb{E}_{p_{\bm \phi}(z|s), \rho_L(s^i)}[D_{\bm \omega}(z)].
        \end{align}
        Notably, while the parameters \({\bm \theta}\) are not explicitly present in the objective, they are implicitly coupled to the objective through the learner's occupancy measure \(\rho_L(s^i)\), which is induced by the learner policy optimized under \(R_{\bm \theta}\). During reward optimization, the abstraction encoder is frozen to prevent representation drift and ensure the reward is optimized over a stable manifold.

        \begin{theorem}[Gradient of reward function]
        \label{theorem:analytic_gradient}
            The analytic gradient of our objective function $\mathcal{L}_\mathcal{F}({\bm \theta})$ presented in Eq. \ref{eqn:trairl-reward-objective} w.r.t ${\bm\theta}$ can be derived as:
            \begin{align} 
                & \nabla_{\bm \theta} \mathcal{L}_\mathcal{F}({\bm \theta}) = \sum\nolimits_{i=1}^n\text{cov}_{\hat{\rho}(s^i),p_{\bm \phi}(z|s)} \left( D_{\bm\omega}(z), \nabla_{\bm \theta} R_{\bm \theta}(z)\right),\nonumber
            \end{align}            
            where $\hat{\rho}(s^i) = \frac{1}{2}(\rho_L(s^i) + \rho_E(s^i))$.
        \end{theorem}

        \begin{figure*}[!ht]
         \centering
         \begin{subfigure}[b]{0.15\textwidth}
             \centering
             \includegraphics[width=\textwidth]{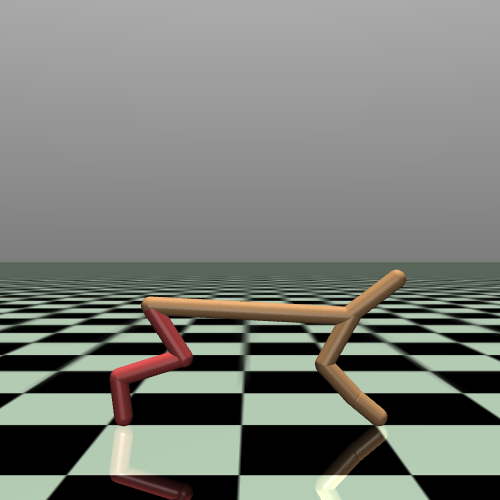}
             \caption{}
             \label{fig:halfcheetah_source_1}
         \end{subfigure}                 
         \begin{subfigure}[b]{0.15\textwidth}
             \centering
             \includegraphics[width=\textwidth]{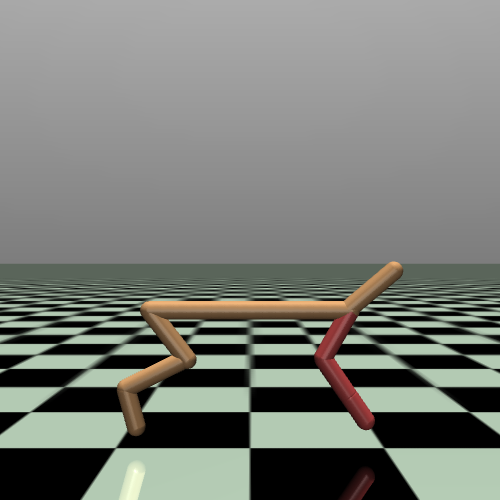}
             \caption{}
             \label{fig:halfcheetah_source_2}
         \end{subfigure}                 
         \begin{subfigure}[b]{0.15\textwidth}
             \centering
             \includegraphics[width=\textwidth]{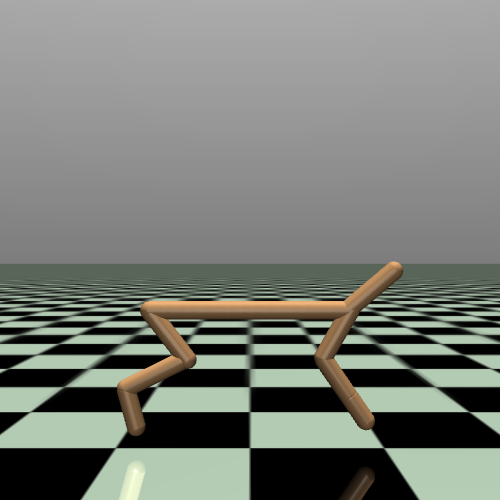}
             \caption{}
             \label{fig:halfcheetah_target}
         \end{subfigure}                 
         \begin{subfigure}[b]{0.15\textwidth}
             \centering
             \includegraphics[width=\textwidth]{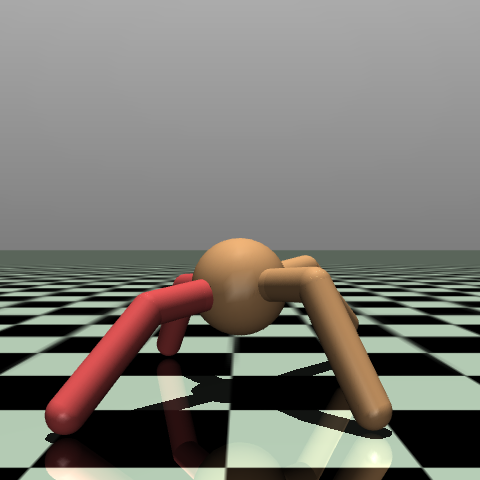}
             \caption{}
             \label{fig:ant_source_1}
         \end{subfigure}                 
         \begin{subfigure}[b]{0.15\textwidth}
             \centering
             \includegraphics[width=\textwidth]{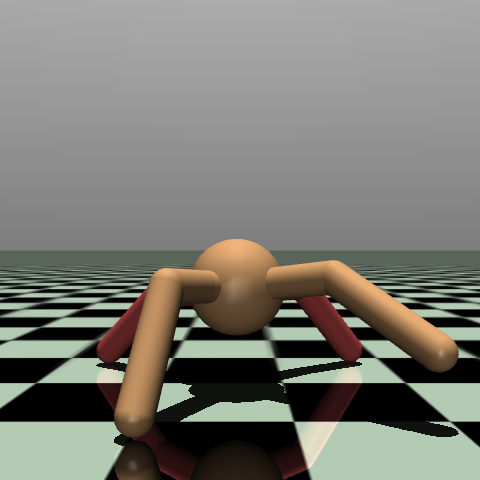}
             \caption{}
             \label{fig:ant_source_2}
         \end{subfigure}                 
         \begin{subfigure}[b]{0.15\textwidth}
             \centering
             \includegraphics[width=\textwidth]{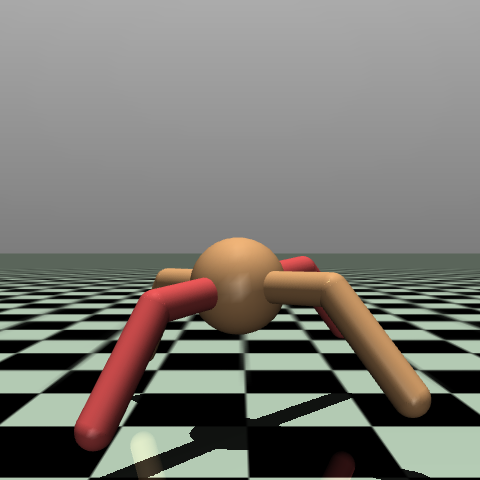}
             \caption{}
             \label{fig:ant_target}
         \end{subfigure}                 
         \caption{Source and target tasks from MuJoCo domains. Red legs are the disabled legs of the robots. Frames (a,b) depict source tasks of running with a disabled leg in {\bf Half Cheetah} while (c) represents the target environment with no disability. Similarly, frames (d,e) show the source tasks of running with different pairs of disabled legs in {\bf Ant}, whereas (f) shows the target of running with another pair of disabled legs.}
         \label{fig:mujoco_source_target_envs}
    \end{figure*}
    
        Theorem~\ref{theorem:analytic_gradient} (proof in Appendix~\ref{appendix:gradient} of supplement) demonstrates that the reward gradient is computed as the covariance between the discriminator's density-matching signal and the reward's parameter sensitivity. This formulation is advantageous as it does not require backpropagation through the environment transition dynamics. During this phase, the encoder \(p_\phi\) is held stationary to prevent representation drift, ensuring the reward is optimized over a stable and continuous manifold. This decoupling prevents the reward from overfitting to task-specific artifacts, thereby enhancing its transferability.
               
        The overall objective function for \tirl{} across $n$ source tasks is a weighted combination of the three objective functions defined previously:
        \begin{align}
            \mathcal{L}({\bm \theta}, {\bm \omega}, {\bm \phi}, {\bm\psi^1},\dots,{\bm\psi^n}) = \lambda_{\text{VAE}}\mathcal{L}_{\text{VAE}}({\bm \phi},{\bm\psi^1},\dots,{\bm\psi^n}) \nonumber\\
            - \lambda_\mathcal{F}\mathcal{L}_\mathcal{F}
            ({\bm \theta})  + \lambda_{\text{WGAN}}\mathcal{L}_{\text{WGAN}}({\bm\phi,\bm\omega})
        \label{eqn:trairl-objective}           
        \end{align}        
        where \(\lambda_{\text{VAE}},\lambda_\mathcal{F}\) and \(\lambda_{\text{WGAN}}\) are the hyperparameters for \(\mathcal{L}_{\text{VAE}}, \mathcal{L}_\mathcal{F}, \mathcal{L}_{\text{WGAN}}\), respectively.

        Finally, \citet{airl} demonstrates that disentangling rewards from task dynamics is essential for transfer, which requires the reward to satisfy a decomposability condition. While this condition is frequently violated in ground-level MDPs due to task-specific transition artifacts, \tirl{} induces an abstracted state manifold designed to satisfy this property. By optimizing \(R_{\bm \theta}(z)\) over a manifold trained for cross-task invariance, \tirl{} aims to recover a reward that is dynamics-agnostic relative to the abstracted transitions. We provide a theoretical analysis and a concrete example in Appendix~\ref{appendix:dynamics-agnostic} demonstrating a case where decomposability fails in the ground MDP but is successfully satisfied within the induced abstract MDP, thereby enabling robust transfer across environments with significantly different dynamics.
        The corresponding algorithm optimizing Eq.~\ref{eqn:trairl-objective} is in Appendix~\ref{appendix:algorithm}.

        \subsection{Analytical Framework for Reward Transferability}
        We aim to formally characterize sufficient conditions under which reward transfer is possible when using \tirl{}, in order to interpret empirical successes and failures.

        \begin{definition}[Reward transferability] 
        Define a reward function $R_{\bm \theta}$ learned for states $\mathcal{S}^i$ of the source task \(\mathcal{T}^i\) as \textbf{transferable} to a target task $\mathcal{T}^t$ iff for a small constant $\epsilon > 0$,
        \begin{align}
            W_1(\rho_{\mathcal{T}^t}^*(z),\rho_{\mathcal{T}^i}(z)) \leq \epsilon \label{eqn:target_expert_target_learner},
        \end{align}    
        where 
        $\rho_{\mathcal{T}^t}^*(z)$ is the abstract state density induced by the (soft-)optimal policy $\pi^*_{\mathcal{T}^t}$ in the target task and $\rho_{\mathcal{T}^i}(z)$ is the abstract state density induced by the policy $\pi_{\mathcal{T}^i}$ obtained by optimizing for $R_{\bm \theta}$ in the source task \(\mathcal{T}^i\).
        \label{def:reward_transferability}
        \end{definition}
        
        Definition \ref{def:reward_transferability} states that the reward function \( R_{\bm \theta} \) is transferable if the policy obtained by optimizing \( R_{\bm \theta} \) in the source task yields an abstract state density that closely matches the one induced by the (soft-)optimal policy in the target task. Next, Theorem~\ref{theorem:trairl-applicability} delineates the conditions under which the reward function learned using \tirl{} is transferable from source tasks to a target task.
        
        \begin{theorem}[Applicability of \tirl{}]
        \label{theorem:trairl-applicability}
        Let $R_{\bm \theta}$ denote the reward function learned by optimizing Eq. $\ref{eqn:trairl-objective}$, $\rho_{\mathcal{T}^i}^*(z)$ denotes the abstract state density of the expert in the $i$-th source task $\mathcal{T}^i$,
        $\epsilon$ be the threshold from Def.~\ref{def:reward_transferability}, and $\alpha\in(0,1)$. If, for every $i$, 
        \begin{align}
            W_1(\rho_{\mathcal{T}^t}^*(z),\rho_{\mathcal{T}^i}^*(z)) & \leq \alpha\epsilon  \label{eqn:similarity_target_source_expert}\\
            W_1(\rho_{\mathcal{T}^i}(z),\rho_{\mathcal{T}^i}^*(z)) & \leq (1-\alpha)\epsilon \label{eqn:similarity_learner_expert_source}
        \end{align}
        then the reward function $R_{\bm \theta}$ is \textbf{transferable} to the target task, enabling effective policy learning.
        \end{theorem}

        \begin{proof}[Sketch of Proof]
              Wasserstein distance, $W_1$, satisfies the \textit{triangle inequality}. Applying it to Eqs.~\(\ref{eqn:similarity_target_source_expert}\) and~\(\ref{eqn:similarity_learner_expert_source}\), we derive Eq.~\(\ref{eqn:target_expert_target_learner}\). This satisfies the transferable reward condition in Def.~\(\ref{def:reward_transferability}\).
        \end{proof}
        
        Eq.~\ref{eqn:similarity_target_source_expert} should be understood as an assumption on the task family, not a condition enforced or verified during training. Despite its simplicity, this decomposition is useful in practice: it explains failure modes (Appendix~\ref{appendix:violation}), motivates the use of abstraction to reduce Wasserstein distances (Table~\ref{table:W1}), and guides the choice of multiple  source tasks.
        Theorem~\ref{theorem:trairl-applicability} identifies two fundamental requirements for cross-task reward transfer. First, Eq.~\ref{eqn:similarity_target_source_expert} represents the {\em structural alignment assumption}: it requires that the optimal behaviors across source and target tasks map to similar regions in the abstracted manifold. This validates the invariance of our latent representation. Second, Eq.~\ref{eqn:similarity_learner_expert_source} represents {\em recoverability correctness}: it ensures that the IRL process effectively distills the source expert's intent into \(R_{\bm \theta}\) such that the resulting learner policy can replicate the expert's occupancy in the abstract space.

    \begin{table*}[!ht]
    \centering
    \small

    \begin{subtable}[ht]{0.59\linewidth}
        \centering        
        \begin{tabular}{c|cc|c}
            \toprule
            & \multicolumn{2}{c}{\textbf{Sources}} & \textbf{Target}\\
            & Run (rear disabled) & Run (front disabled) & Run (no disability) \\
            \midrule
            \firl & 3,252.72 $\pm$ 67.9 & 3,523.01 $\pm$ 91.1 & 3,582.10 $\pm$ 94.3 \\
            AIRL-ME & 4,014.52 $\pm$ 79.3 & 3,905.38 $\pm$ 73.1 & 3,725.63 $\pm$ 80.9 \\
            RIME  & 4,271.67 $\pm$ 36.1 & 4,129.19 $\pm$ 83.4 & 4,061.22 $\pm$ 58.6 \\
            I2L   & 4,396.43 $\pm$ 45.4 & 4,518.66 $\pm$ 52.2 & 4,512.71 $\pm$ 66.4 \\
            \textbf{\tirl{}} & 4,404.07 $\pm$ 57.6 & 4,359.35 $\pm$ 99.2 & \textbf{5,835.11} $\pm$ \textbf{74.0} \\
            \midrule
            Expert & 5,052.25 $\pm$ 25.4 & 5,499.07 $\pm$ 156.1 & 6,420.38 $\pm$ 37.9 \\
            \bottomrule        
        \end{tabular}
        \caption{}
        \label{table:halfcheetah}
    \end{subtable}    
    \hfill
    \begin{subtable}[ht]{0.39\linewidth}
    \centering
    \begin{tabular}{@{}llcc@{}}
        \toprule
        Domain & Pair & Abstraction & Ground \\
        \midrule
        \multirow{3}{*}{HalfCheetah}
        & S1--S2 & \textbf{0.36} & 1.37 \\
        & S1--T  & \textbf{0.62} & 2.83 \\
        & S2--T  & \textbf{0.55} & 2.10 \\
        \midrule
        \multirow{3}{*}{Ant}
        & S1--S2 & \textbf{0.33} & 1.78 \\
        & S1--T1 & \textbf{0.71} & 2.82 \\
        & S1--T2 & \textbf{0.79} & 2.97 \\
        \bottomrule
    \end{tabular}    
    \caption{}
    \label{table:W1}
    \end{subtable}
    \caption{Transfer performance and abstraction analysis. 
        (a) Mean cumulative rewards with standard deviation for HalfCheetah. 
        (b) Cross-task Wasserstein distances \(W_1\), where S1 and S2 denote source tasks, and T, T1, and T2 denote target tasks.}
\end{table*}

\begin{table*}[!ht]
            \centering
            \small       
            \setlength\tabcolsep{2.5pt}
            
            \begin{tabular}{c|cc|ccc}
                \toprule
                & \multicolumn{2}{c}{\bf Sources} & \multicolumn{3}{c}{\bf Targets}\\
                & Leg 1,2 disabled & Leg 0,3 disabled & Leg 1,3 disabled & Leg 0,2 disabled & Half Cheetah (One-Shot) \\ 
                \midrule
                \firl & 2,456.17 $\pm$ 85.0 & 1,146.39 $\pm$ 95.8 & 1,598.24 $\pm$ 44.9 &  1,652.89 \(\pm\) 74.3 & 3871.57 \(\pm\) 115.1\\
                AIRL-ME & 2,389.65 $\pm$ 52.0 & 2,231.09 $\pm$ 87.6 & 2,098.11 $\pm$ 99.3 & 2,190.12 $\pm$ 50.2 & 4963.55 $\pm$ 215.7 \\
                RIME  & 2,681.67 $\pm$ 49.9 & 2,708.14 $\pm$ 81.5 & 2,190.71 $\pm$ 61.9 & 2,188.00 $\pm$ 59.8 & 4623.12 $\pm$ 155.0 \\
                I2L   & 2,831.28 $\pm$ 36.4 & 2,786.90 $\pm$ 79.4 & 2,585.32 $\pm$ 84.5 & 2,618.32 \(\pm\) 92.4 & 4789.89 $\pm$ 90.4\\
                \bf{\tirl{}}  & 2,714.18 $\pm$ 35.9 & 2,936.52 $\pm$ 95.5 & \textbf{2,917.92} $\pm$ \textbf{79.3} & \textbf{3,156.54} \(\pm\) \textbf{63.1} & \textbf{5,378.78} \(\pm\) \textbf{61.7} \\
                \midrule
                Expert  & 3,312.12 $\pm$ 304.3 & 3,303.99 $\pm$ 341.0 & 3,369.05 $\pm$ 216.8 & 3,590.57 \(\pm\) 158.2 & 6,420.38 $\pm$ 37.9 \\
                \bottomrule
            \end{tabular}
            \caption{Mean cumulative rewards with standard deviation for \textbf{Ant}. \tirl{} achieves the highest cumulative rewards in all the target tasks. Comprehensive experiments including more source and target tasks can be found in Appendix~\ref{appendix:comprehensive_experiments}.}
            \label{table:tirl_ant_result}
        \end{table*}

\section{Experiments}
\label{sec:experiments}

    We implement our framework in PyTorch and evaluate its performance across MuJoCo~\citep{mujoco} and Assistive Gym~\citep{DBLP:journals/corr/abs-1910-04700} benchmarks. \tirl{} is trained to convergence using 50 trajectories from multiple source tasks with distinct dynamics. 
    Policy performance is measured by the mean and standard deviation of accumulated rewards over 25 evaluation episodes.
    For a fair comparison, we benchmark against three state-of-the-art techniques: AIRL-ME~\citep{airl-me}, RIME~\citep{rime}, and I2L. While RIME and I2L are specifically designed for dynamics shifts, they lack the abstraction mechanisms utilized by \tirl{}. All algorithms receive the default Gymnasium observation space, including joint angles and velocities, as input.
    
    \paragraph{Model architecture and implementation.} We employ a multilayer perceptron with Tanh as the activation function for both the encoder and decoder in VAE and to represent the reward function. We use Soft Actor-Critic (SAC)~\citep{sac} as the backbone RL algorithm. Further details regarding the model architecture and hyperparameters to aid reproducibility are available in Appendix \ref{appendix:training}. 

    \subsection{Evaluations in MuJoCo-Gym} \label{sec:experiments-tirl}
    
    We use the {\bf Half Cheetah} and {\bf Ant} domains in MuJoCo-Gym. Figure~\ref{fig:mujoco_source_target_envs} illustrates the source and target tasks, which differ in dynamics between the sources and between the sources and the target. Specifically, these differences in dynamics arise from {\em disabling} different pairs of legs. Disabled legs are indicated in red in the frames of Fig.~\ref{fig:mujoco_source_target_envs}. While the action space remains unchanged across the environments, as the input actions are directly applied to all joints regardless of whether a leg is disabled, the dynamics differ because the disabled legs cannot respond to the input actions.

    With a single source task, abstraction learning is not well defined and tends to entangle reward-relevant and dynamics-specific features (Appendix~\ref{appendix:single_source}). Multiple  sources allow for invariance that cannot be explained by dynamics alone, enabling the disentanglement of task-invariant behavioral structure. In the target task, policies are trained exclusively using the learned reward function without access to the environment's ground-truth reward or expert demonstrations. For evaluation only, we report performance using the environment’s true reward function, following standard practice in IRL benchmarks.
        
        \begin{figure*}[!ht]
        \centering
            \includegraphics[width=0.8\linewidth]{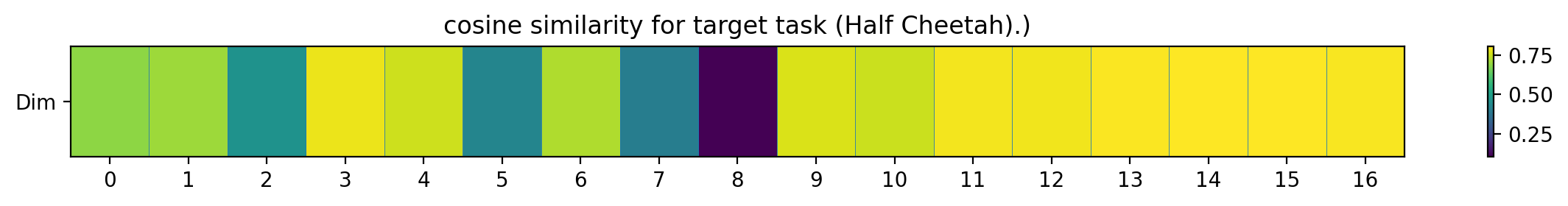}
            \caption{Visualization of abstraction sensitivity on the HalfCheetah target task. The x-axis indexes ground-state dimensions, and darker colors indicate lower similarity, meaning the abstraction is more sensitive to that dimension.}
            \label{fig:semantic_analysis_cosine_similarity_main_paper}
        \end{figure*}
        
    \paragraph{Reward Transferability.}
        Tables~\ref{table:halfcheetah} and~\ref{table:tirl_ant_result} report our results using \tirl{} on the baselines in Half Cheetah and Ant domains, respectively. These show the mean cumulative rewards obtained in the target environment as well as in the two source environments. The optimal policy's (expert) rewards for each are reported as well.         
        Observe that the rewards learned by \tirl{} achieve the highest average return compared to all baselines in the target tasks for both domains. As such, the abstracted rewards learned by \tirl{} from the two sources are most transferable and correct. I2L achieves the next best performance on the target task but remains significantly lower than \tirl{}'s (Student's paired t-test, $p < 0.01$). More detailed experiments, including 10 source and 5 target tasks, are provided in Appendix \ref{appendix:comprehensive_experiments}. Also refer to Appendix~\ref{appendix:violation} for the failure case of \tirl{} when the structural alignment assumption (Eq.~\ref{eqn:similarity_target_source_expert}) is violated.

        We evaluate \tirl{} in a more challenging cross-domain setting: transfer from Ant to the Half Cheetah domain. We empirically demonstrate the transferability between domains with different dynamics and states of inversely learned rewards. Despite these differences, \tirl{} shows strong performance in the target task with one-shot transfer, as shown in the last column of Table~\ref{table:tirl_ant_result}. Details of this experiment are provided in Appendix \ref{appendix:one-shot}. Furthermore, Appendix~\ref{appendix:semantic_analysis} provides a semantic analysis of the abstract state space showing that an abstraction trained on quadrupeds captures transferable structure useful for bipedal locomotion, rather than merely encoding peculiarities of the expert data such as joint angles.

       \begin{figure*}[!ht]
            \centering            
            \begin{subfigure}[b]{0.22\linewidth}
             \includegraphics[width=\linewidth]{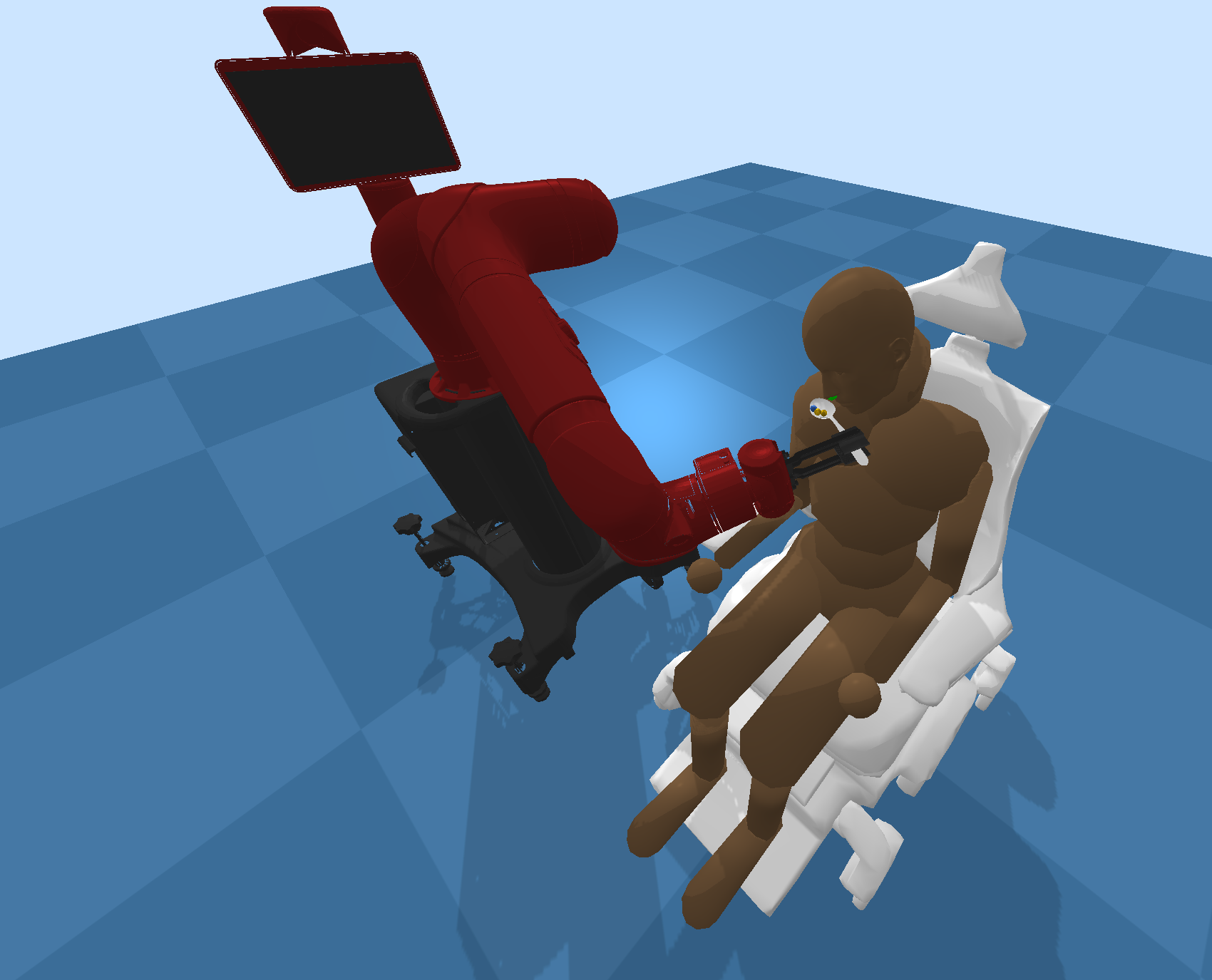}
             \caption{\small FeedingSawyer}
             \label{fig:feeding_goal}
            \end{subfigure}
            \begin{subfigure}[b]{0.22\linewidth}
             \centering
             \includegraphics[width=\linewidth]{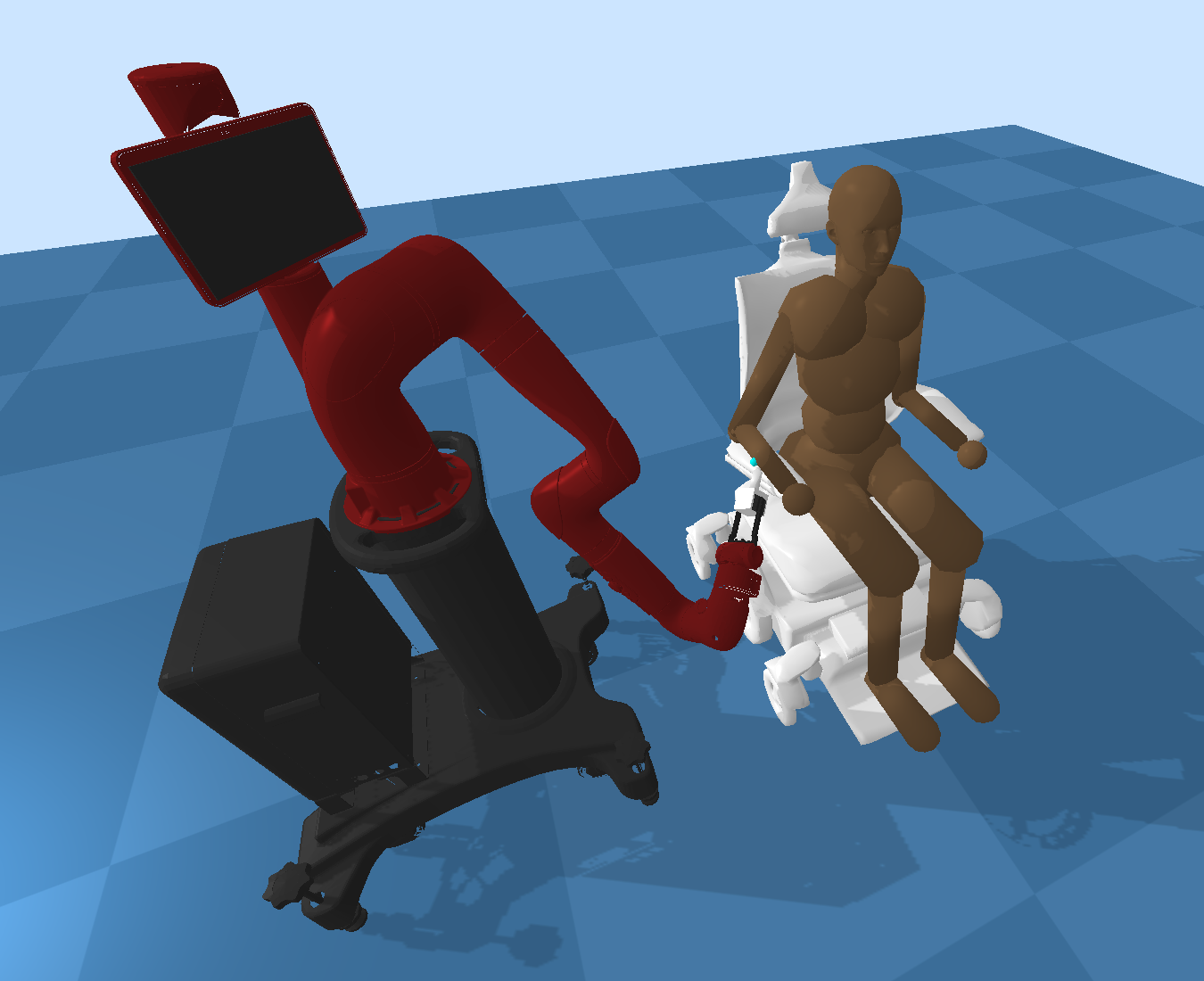}
             \caption{\small ScratchItchSawyer}
             \label{fig:scratch_itch_goal}
            \end{subfigure}
            \begin{subfigure}[b]{0.55\linewidth}
            \centering
            \begin{small}
            \setlength\tabcolsep{2pt}
            \begin{tabular}{c|ccc} 
                \toprule
                & \multicolumn{2}{c}{\bf Sources} & \textbf{Target}\\
                & Feeding task 1 & Feeding task 2 & Scratch itch\\
                \midrule
                \firl &  -10.63 $\pm$ 16.79 &  -15.3 $\pm$ 22.78 & -21.35 $\pm$ 14.89 \\
                AIRL-ME & -6.35 $\pm$ 25.1 & 3.59 $\pm$ 17.3 & -17.32 $\pm$ 11.6 \\
                RIME    &  8.01 $\pm$ 6.0 & 8.66 $\pm$ 9.7 & -11.64 $\pm$ 4.2 \\
                I2L   &    9.32 $\pm$ 11.8 &    9.11 $\pm$ 11.4 & -10.07 $\pm$ 8.0 \\
                \bf{\tirl{}} & 8.32 $\pm$ 7.9 & 9.56 $\pm$ 10.5 & \textbf{-3.82} $\pm$ \textbf{3.3}\\
                \midrule
                Expert & 11.29 $\pm$ 5.3 & 12.77 $\pm$ 4.2 & -1.18 $\pm$ 5.7 \\
                \bottomrule
            \end{tabular}
            \caption{}
            \label{table:tirl_assistive_gym_result}
            \end{small}            
            \end{subfigure}
            \caption{Two tasks in the feeding domain of \textbf{Assistive Gym} (a) are used as sources to learn a reward function that is transferred to perform the task of scratching an itch shown in (b). (c) Learned policy returns with standard deviation in the \textbf{Assistive Gym} environments. \tirl{} shows the highest return in the target task.}
            \label{fig:tirl_assistive_gym_result}
        \end{figure*} 

        \paragraph{Benefit of Abstraction.} \label{sec:benefit_of-abstraction}
        To understand why \tirl{} performs better, we aim to gain some insight into our novel abstraction concept.
        In Table~\ref{table:W1}, we report the 1-Wasserstein distance (\(W_1\)) between the abstracted state densities of the two source tasks and between the densities of each source and the target, \(W_1(\rho^*_{\mathcal{T}^i}(z), \rho^*_{\mathcal{T}^t}(z))\), for the Half Cheetah and Ant. We compare these to the corresponding $W_1$ distances between the ground state densities.  To obtain the \(W_1\) between ground state densities, we train a variant of $f$-IRL (see Appendix A.2 of \citet{firl}). 
        Although expert trajectories for the target are not available in the core transfer setting, the Wasserstein distances between abstract expert occupancies of source tasks (Table \ref{table:W1}) serve as an empirical proxy for alignment. When these distances are large, we observe degraded transfer performance (Appendix \ref{appendix:violation}).

        Next, we perform an abstraction sensitivity analysis to investigate the underlying mechanisms of the learned abstraction. The details of the procedure and comprehensive results are provided in Appendix~\ref{appendix:extra experiments}. As illustrated in Figure~\ref{fig:semantic_analysis_cosine_similarity_main_paper}, the abstract representation exhibits the highest sensitivity to the $x$-velocity of the front tip. This finding is intuitive as this dimension directly captures the robot's directional progress and velocity, aligning with the forward-speed component of the ground-truth reward function. The fact that the encoder prioritizes this feature suggests that the abstraction process effectively distills task-relevant dynamics while maintaining invariance to less critical state dimensions.

        For each comparison in Table~\ref{table:W1}, the abstracted state densities yield a consistently lower \(W_1\) distance compared to the ground state densities. This indicates that the abstraction reduces task-specific variability in the abstract state space, bringing the source and target closer in terms of their occupancy measures. In particular, the \(W_1\) distance between the abstract state densities of optimal policies in the source and target tasks, \(W_1(\rho^*_{\mathcal{T}^i}(z), \rho^*_{\mathcal{T}^t}(z))\), corresponds to Eq.~\ref{eqn:similarity_target_source_expert} in Theorem~\ref{theorem:trairl-applicability}. A smaller distance implies a tighter bound on the transfer error \(\epsilon\) and therefore reflects better generalization of the abstract state space across tasks.

        We explored the sensitivity of \tirl{}'s performance to hyperparameter values and conducted ablation studies. The results are reported in Appendix \ref{appendix:hyperparameter_sensitivity}. The visualizations of the abstracted state manifolds are reported in Appendix~\ref{appendix:tsne}. 
    
    \subsection{Evaluation in Human-Robot Assistive Gym} \label{sec:assistive_gym}

    To give an indication of the utility of \tirl{} in the real-world, we illustrate its use in human-robot collaboration using the  realistic Assistive Gym testbed~\citep{DBLP:journals/corr/abs-1910-04700}. Similar to MuJoCo, Assistive Gym is a physics-based simulation framework but for studying human-robot interaction and robotic assistance by the collaborative robot Sawyer.

        For the source tasks, we select {\em FeedingSawyer}, a simulation environment in which the collaborative robot is tasked with feeding a disabled human (Fig.~\ref{fig:feeding_goal}) The two source tasks differ in the condition of the disabled human. The human is static in one while the human has tremors in the other, which cause the target area to move leading to a shifting goal. The robot's challenge is to adapt to these differing human conditions while executing the feeding task. Our target task is a different task, {\em ScratchItchSawyer}, in which Sawyer is tasked with scratching a disabled human's itch (Fig.~\ref{fig:scratch_itch_goal}). Although this task differs from feeding in terms of its specific goal, it is also aligned in that the robot must move its end effector precisely to a designated target area. This abstract information should be represented by the learned reward function.        

        We report the performance of \tirl{} and the baselines in transferring the reward function inversely learned from the two feeding tasks, to learn how to scratch the person's itch. Table~\ref{table:tirl_assistive_gym_result} gives the mean cumulative reward from forward RL in the target using the learned rewards. Notice that \tirl{} yields the policy with the highest value and close to the optimal, indicating transferable rewards. The transferred reward function exhibits a strong positive linear relationship with the ground-truth rewards, as indicated by a Pearson correlation coefficient of 0.86 ($p < .001$). 
        Among the baselines, I2L and RIME achieve comparative transfer, with performance close to each other on the source tasks but still weaker on the target. AIRL-ME lags further behind, showing limited ability to generalize. In contrast, \tirl{} consistently outperforms all three, indicating that the abstracted reward function provides superior transferability.

\section{Conclusion}

    \tirl{} represents a significant advancement of IRL by introducing a principled approach to inversely learn transferable reward functions from demonstrations of multiple aligned tasks. The key contribution lies in its ability to extract invariant abstractions that model the structure intrinsic to multiple tasks, which makes them transferable to aligned target tasks. \tirl{}'s analytical properties delineate the transfer applicability of the abstracted rewards and the experiments validate the transferability in both formative and use-inspired contexts. By explicitly leveraging multiple  source tasks, \tirl{} mitigates the entanglement between task-specific dynamics and reward-relevant structure, enabling more robust generalization under dynamics shift. 

\section*{Acknowledgments}

This research was supported in part by NSF grants IIS-1830421 and IIS-2312657 both to Doshi. We acknowledge useful discussions with Prasanth Suresh that informed parts of this paper.



\onecolumn
\newpage
\appendix 
\addcontentsline{toc}{section}{Appendix} 
\part{Appendix} 
\parttoc 

\newpage

\section{Algorithm} \label{appendix:algorithm}
\textbf{Algorithm $\ref{algo:tirl-training}$ demonstrates the training procedure for \tirl{}}. During each iteration, trajectories are uniformly sampled from each source environment and policy, and added to the buffer (lines~$\ref{algo_step:add_trajecotries}$-$\ref{algo_step:sample}$). The encoder $p_{\bm\phi}$, decoders $q_{\bm\psi^1}, ..., q_{\bm\psi^n}$, discriminator $D_{\bm \omega}$, and the reward function $\mathcal{R}_{\bm \theta}$ are updated (line~$\ref{algo_step:update_trairl}$). Then, the learner policies $\pi_{\bm \xi^1}, ..., \pi_{\bm\xi^n}$ are updated in the source tasks $\mathcal{T}^1, ..., \mathcal{T}^n$ using the updated reward function $R_{\bm \theta}$, respectively (line~$\ref{algo_step:update_policy}$).

        \begin{algorithm}
            \small
            \caption{\tirl{}: \textit{Training Phase}} 
            \begin{algorithmic}[1]
                    \Require Expert trajectories $\tau^1_E, ..., \tau^n_E$; Source tasks: $\mathcal{T}^1, ..., \mathcal{T}^n$
                    \State Initialize learner policies $\pi_{\bm\xi^1}, ..., \pi_{\bm\xi^n}$; Trajectory buffer $B$; Discriminator $D_{\bm\omega}$; Reward function $R_{\bm\theta}$; encoder $p_{\bm \phi}$; and decoders $q_{\bm\psi^1}, ..., q_{\bm\psi^n}$
                    \State Add expert trajectories $\tau^1_E, ..., \tau^n_E$ into trajectory buffer $B$
                    \While {$\pi_{\xi^1}, ..., \pi_{\xi^n}$ continue improving within $k$ steps}
                        \For {task $i$ in ${1, ... ,n}$}
                            \State Collect state-only trajectories $\tau^i = (s_0, ..., s_T)$ \label{algo_step:collect_trajecotries}
                            \State Add trajectories $\tau^i$ into trajectory buffer $B$ \label{algo_step:add_trajecotries}
                        \EndFor
                        \State Uniformly sample trajectories $\tau$ from Buffer $B$ \label{algo_step:sample}
                        \State Update $R_{\bm \theta}$; $D_{\bm \omega}$; $p_{\bm\phi}$; and $q_{\bm\psi^1}, ..., q_{\bm\psi^n}$ using $\tau$ by Eq.~$\ref{eqn:trairl-objective}$ \label{algo_step:update_trairl}

                        \State Update $\pi_{\bm \xi^1}, ..., \pi_{\bm\xi^n}$ using $R_{\bm \theta}$ \label{algo_step:update_policy}
                    \EndWhile
            \end{algorithmic} 
                \label{algo:tirl-training}
        \end{algorithm}

        \begin{algorithm}
        \small
        	\caption{\tirl{}: \textit{Transfer Testing Phase}} 
        	\begin{algorithmic}[1]
                    \Require Reward function $R_{\theta}$ learned by Algorithm $\ref{algo:tirl-training}$, Target task $\mathcal{T}^t$
                    \State Initialize policy $\pi_{\xi}$
                    \While {$\pi_\xi$ continues improving within $k$ steps}
                        \State Update $\pi_{\xi}$ only using the learned reward function $R_{\theta}$ in target task $\mathcal{T}$
                    \EndWhile
        	\end{algorithmic} 
                 \label{algo:tirl-testing}       
         \end{algorithm}

\section{Proof of Theorems}
\subsection{Analytical Gradient of \tirl{} (Theorem \ref{theorem:analytic_gradient})} \label{appendix:gradient}
    \begin{proof}
    In this section, we derive the analytic gradient of the proposed \tirl{} (Theorem \ref{theorem:analytic_gradient}). From \citep{firl}, we have the following equations:
    \begin{small}\begin{equation}
        \label{eqn:rho_theta_s gradient}
        \begin{split}
            \frac{d\rho_L(s)}{d\theta} & = \int \frac{d\rho_L(s)}{dR_\theta(s^*)}\frac{dR_\theta(s^*)}{d\theta}ds^* \\
                                           & = \frac{1}{Z}\int p(\tau) e^{\sum^{T}_{t=1} R_\theta(s_t)}\eta_\tau(s)\sum^T_{t=1}\frac{dR_\theta(s_t)}{d\theta}d\tau -T\rho_L(s)\int \rho_L(s^*)\frac{dR_\theta(s^*)}{d\theta}ds^*,
        \end{split}
    \end{equation}
    \end{small}
    where $Z$ is the normalization constant, and \(\eta_\pi(s)\) denotes the number of times a state occurs in a trajectory \(\tau\).
    
    The joint distribution over states and abstracted states is defined as:
    \begin{equation}
        \begin{split}
            \label{eqn:z_s_joint_dist}
            p(z, s) & = p_{\bm \phi}(z|s)\rho(s), \\
        \end{split}
    \end{equation}
    where $p_{\bm \phi}(z|s)$ is the encoder parameterized by ${\bm \phi}$.
    
    The marginal distribution of the abstraction $z$, denoted as the abstract state density $\rho(z)$, is obtained by integrating out the state $s$ from Eq.~\ref{eqn:z_s_joint_dist}:    
    \begin{align*}
        \rho(z) & = \int_\mathcal{S} p(z, s) ds \\
                & = \int_\mathcal{S} p_{\bm \phi}(z|s)\rho(s) ds.
    \end{align*}

    When operating in the abstracted state space, Eq.~\ref{eqn:rho_theta_s gradient} becomes:
    \begin{small}\begin{equation}
        \label{eqn:rho_theta_z gradient}
        \begin{split}
            \frac{d\rho_L(z)}{d\theta} & = \int \frac{d\rho_L(z)}{dR_\theta(z^*)}\frac{dR_\theta(z^*)}{d\theta}dz^* \\
                                           & = \frac{1}{Z}\int p(\tau) e^{\sum^{T}_{t=1} R_\theta(z_t)}\eta_\tau(z)\sum^T_{t=1}\frac{dR_\theta(z_t)}{d\theta}d\tau -T\rho_L(z)\int \rho_L(z^*)\frac{dR_\theta(z^*)}{d\theta}dz^*,
        \end{split}
    \end{equation}
    \end{small}
    
    When optimizing the abstract state density matching objective between the expert density $\rho_E(z)$ and the learner density $\rho_L(z)$ in the $i$-th source environment, we measure their discrepancy using 1-Wasserstein distance. The objective is formulated as:
    \begin{align*}
            \mathcal{L}_\mathcal{F}({\bm \theta}) & = \sum_{i=1}^n W_1(\rho_E(z) , \rho_L(z)) \\
            & = \sum_{i=1}^n \sup_{||f||_L\leq 1} \left| \mathbb{E}_{z\sim\rho_E(z)}[f(z)] - \mathbb{E}_{z\sim\rho_L (z)}[f(z)] \right| \\
            & = \sum_{i=1}^n \max_{D_{\bm \omega}} \mathbb{E}_{z\sim\rho_E(z)}[D_{\bm \omega}(z)] - \mathbb{E}_{z\sim\rho_L(z)}[D_{\bm \omega}(z)] \\
            & = \sum_{i=1}^n \max_{D_{\bm \omega}} \int_\mathcal{\bar{S}} D_{\bm \omega}(z)\rho_E(z) dz - \int_\mathcal{\bar{S}} D_{\bm \omega}(z)\rho_L(z) dz.
    \end{align*}
    The objective is derived using the Kantorovich–Rubinstein duality, which reformulates the 1-Wasserstein distance as a supremum over all 1-Lipschitz functions. To approximate this function, we introduce a discriminator $D_{\bm \omega}(z)$ and express the optimization as a maximization of the expected difference between expert and learner distributions. To ensure that $D_{\bm \omega}(z)$ satisfies the 1-Lipschitz constraint required by the duality, we apply a gradient penalty, which also stabilizes optimization while preserving theoretical correctness.

    The gradient of the objective, Eq.~\ref{eqn:trairl-reward-objective}, w.r.t $\bm\theta$ is derived as:
    \begin{small}
    \begin{equation}
        \begin{split}
            \nabla_{\bm\theta} \mathcal{L}_\mathcal{F}({\bm \theta}) 
            & = - \sum_{i=1}^n \int_\mathcal{\bar{S}} D_{\bm \omega}(z)\nabla_{\bm\theta}\rho_L(z) dz 
        \end{split}
    \end{equation}
    \end{small}

    Substituting the gradient of abstract state density $\rho_L(z)$ w.r.t $\theta$ with Eq.\ref{eqn:rho_theta_z gradient}, we have:
    \begin{small}
    \begin{align}
            \nabla_{\bm\theta} \mathcal{L}_\mathcal{F}({\bm \theta}) 
            & \propto \frac{1}{T} \sum_{i=1}^n \int \rho_L(\tau^i) \sum^T_{t=1} D_{\bm \omega}(z_t) \sum^T_{t=1}\frac{dR_{\bm\theta}(z_t)}{d{\bm\theta}} d\tau^i \nonumber\\
            & ~ - T ~\sum_{i=1}^n ~\int_\mathcal{\bar{S}} D_{\bm\omega}(z)\rho_L(z)\left( \int_\mathcal{\bar{S}} \rho_L(z^*)\frac{dR_\theta(z^*)}{d{\bm\theta}} dz^*\right)dz \nonumber\\
            & = \frac{1}{T} \sum_{i=1}^n \mathbb{E}_{\tau\sim \rho_L(\tau^i)}\left[\sum^T_{t=1} D_{\bm \omega}(z) \right] \sum^T_{t=1}\frac{dR_{\bm\theta}(z_t)}{d{\bm\theta}}  \nonumber\\
            & ~ - T ~\sum_{i=1}^n ~\mathbb{E}_{z \sim\rho_L(z)}\left[D_{\bm \omega}(z)\right]\mathbb{E}_{z \sim\rho_L(z)}\left[\frac{dR_{\bm\theta}(z)}{d{\bm\theta}}\right].
        \end{align}
    \end{small}
    
    To gain further intuition about this equation, we can express all the expectations in terms of trajectories:   
    \begin{small}
    \begin{align}
            \nabla_{\bm\theta} \mathcal{L}_\mathcal{F}({\bm \theta}) 
            & \propto \frac{1}{T}\sum_{i=1}^n \Bigg( \sum^T_{t=1} D_{\bm \omega}(z_t) \sum^T_{t=1}\nabla_{\bm\theta}R_{\bm\theta}(z_t)  \nonumber\\
            & ~~~~~- \left. \mathbb{E}_{\rho_L(\tau^i)}\left[\sum^T_{t=1} D_{\bm \omega}(z_t) \right]\mathbb{E}_{\rho_L(\tau^i)}\left[\sum^T_{t=1}\nabla_{\bm\theta}R_{\bm\theta}(z_t)\right]\right) \nonumber\\
            & \propto \sum_{i=1}^n \sum^T_{t=1} \text{cov}_{\tau\sim\rho_L(\tau^i)} \Bigg( D_{\bm \omega}(z_t), \nabla_{\bm\theta} R_{\bm\theta}(z_t)\Bigg). \nonumber\\
            & = \sum_{i=1}^n \text{cov}_{z\sim\rho_L(z)} \Bigg( D_{\bm \omega}(z), \nabla_{\bm\theta} R_{\bm\theta}(z)\Bigg). \nonumber\\
            & = \sum_{i=1}^n\text{cov}_{s\sim\rho_L(s^i),z\sim p_{\bm \phi}(z|s)} \left( D_{\bm\omega}(z), \nabla_{\bm \theta} R_{\bm \theta}(z)\right).\label{eqn:tirl_origin_gradeint}
    \end{align}
    \end{small}

    When operating in high-dimensional observation domains, a significant impediment arises if the state visitation distributions induced by the learner's current policy substantially diverge from the expert demonstration trajectories. Under such conditions of distributional mismatch, we empirically find that the gradient signal derived from our proposed objective function, Eq.~\ref{eqn:tirl_origin_gradeint}, provides limited supervisory information to guide the policy optimization process. We follow the technique introduced by \citep{finn2016guided}, mixing the data samples from expert trajectories with the learner trajectories. The revised objective function is given in the following.
    \begin{small}
        \begin{equation}        
            \label{eqn:tirl-reward-practical}
            \nabla_{\bm\theta} \mathcal{L}_\mathcal{F}({\bm \theta}) = \sum_{i=1}^n\text{cov}_{s\sim \hat{\rho}(s^i),z\sim p_{\bm \phi}(z|s)} \left( D_{\bm\omega}(z), \nabla_{\bm \theta} R_{\bm \theta}(z)\right),
        \end{equation}
        \end{small}
        where $\hat{\rho}(s^i) = \frac{1}{2}(\rho_L(s^i) + \rho_E(s^i))$.
    \end{proof}
\subsection{Proof of Theorem 2}
    \begin{proof}
        We begin by noting that the 1-Wasserstein distance is a \textbf{metric} and therefore satisfies the triangle inequality: \begin{align}
            W_1(\rho_1, \rho_3)\leq W_1(\rho_1, \rho_2) + W_1(\rho_2, \rho_3). \nonumber
        \end{align}
        Apply the triangle inequality to the three distributions:
        \begin{itemize}
            \item \(\rho^*_{\mathcal{T}^t}(z)\) - abstract state density of the expert in the target task.
            \item \(\rho^*_{\mathcal{T}^i}(z)\) - abstract state density of the expert in the \(i\)-th source task.
            \item \(\rho_{\mathcal{T}^i}(z)\) - abstract state density of the learner in the \(i\)-th source task.
        \end{itemize}
        By the triangle inequality:
        \begin{align*}
            W_1(\rho^*_{\mathcal{T}^t}(z), \rho_{\mathcal{T}^i}(z)) 
            & \leq W_1(\rho^*_{\mathcal{T}^t}(z), \rho^*_{\mathcal{T}^i}(z)) + W_1(\rho^*_{\mathcal{T}^i}(z), \rho_{\mathcal{T}^i}(z)) \\  
            & \leq \alpha\epsilon + (1-\alpha)\epsilon \\
            & = \epsilon.
        \end{align*}
        This satisfies the condition in Def~\ref{def:reward_transferability}, which requires that the abstract state density induced by the learned reward function \(R_{\bm\theta}\) in a source task is close (within \(\epsilon\)) to the optimal policy's abstract state density in the target task.
        
        Hence, \(R_{\bm\theta}\) is transferable.
    \end{proof}

    \subsection{Dynamics Disentangled State-Only Reward Function} \label{appendix:dynamics-agnostic}
    In this section, we follow the derivations and definitions of \cite{airl} to establish that \tirl{} learns disentangled reward functions. For completeness, we restate the key definitions and theorems here. We first define the induced ground-level reward function using the abstract reward function in \tirl{}.

    \begin{definition} [Induced ground-level reward function]
       Let \(\phi:\mathcal{S}\rightarrow\mathcal{Z}\) be \tirl{}'s abstraction function and let \(r_{\text{abs}}:\mathcal{Z}\rightarrow\mathbb{R}\) be \tirl{}'s abstract reward function. Define the induced ground-level reward function
       $$r_\phi(s)\coloneq r_{\text{abs}}(\phi(s)).$$
    \end{definition}

    Then we borrow the definition of "disentangled rewards" from \cite{airl}.
    
    \begin{definition} [Disentangled rewards]
        A reward function \(r'(s,a,s')\) is (perfectly) disentangled with respect to a ground-truth reward \(r(s,a,s')\) and a set of dynamics \(\mathcal{T}\) such that under all dynamics \(T\in\mathcal{T}\), the optimal policy is the same: \(\pi^*_{r',T}(a\mid s)=\pi^*_{r,T}(a\mid s)\).
    \end{definition}

    Disentangled rewards can be informally understood as reward functions that induce the same optimal policy as the ground truth reward under any admissible dynamics. To demonstrate how \tirl{} recovers such a reward, we first recall the definition of the decomposability condition.

    \begin{definition} [Decomposability condition]
        Two states \(s_1,s_2\) are defined as "1-step linked" under a dynamics or transition distribution \(T(s'\mid a,s)\) if there exists a state \(s\) that can reach \(s_1\) and \(s_2\) with positive probability in one time step. Also, we define that this relationship can transfer through transitivity: if \(s_1\) and \(s_2\) are linked, and \(s_2\) and \(s_3\) are linked, then we also consider \(s_1\) and \(s_3\) to be linked.\\
        A transition distribution \(T\) satisfies the decomposability condition if all states in the MDP are linked with all other states.
    \end{definition}

    Theorem \ref{theorem:airl-1} and \ref{theorem:airl-2} formalize that \tirl{} recovers reward functions disentangled from the dynamics.
    
    \begin{theorem} \label{theorem:airl-1}
        Let \(r(s)\) be a ground-truth reward, and \(T\) be a dynamics model satisfying the decomposability condition. Suppose IRL recovers a state-only reward \(r'(s)\) such that it produces an optimal policy in \(T\):
        $$Q^*_{r',T}(s,a)=Q^*_{r,T}(s,a)-f(s).$$
        Then, \(r'(s)\) is disentangled with respect to all dynamics.
    \end{theorem}
    \begin{proof}
        Refer to Theorem 5.1 in \cite{airl}.
    \end{proof}

    \begin{theorem} \label{theorem:airl-2}
        If a reward function \(r'(s,a,s')\) is disentangled for all dynamics functions, then it must be state-only, i.e. if for all dynamics \(T\),
        $$Q^*_{r,T}(s,a)=Q^*_{r',T}(s,a)+f(s) \hspace{0.1in}\forall s,a.$$
        Then \(r'\) is only a function of state.
    \end{theorem}
    \begin{proof}
        Refer to Theorem 5.2 in \cite{airl}.
    \end{proof}

    Next, we demonstrate an example where the decomposability condition is not satisfied, whereas in the \tirl{}, a disentangled reward function can still be learned. Consider the following 3-state MDP with deterministic dynamics and starting state A:
    
    \begin{center}
        \begin{tikzpicture}[
          >=Stealth,
          state/.style={circle, draw, minimum size=18pt, inner sep=0pt, font=\small}
        ]
            \node[state] (A) {A};
            \node[state, right=2.2cm of A] (B) {B};
            \node[state, right=2.2cm of B] (C) {C};
            
            \draw[->] (A) -- node[midway, above]{b, 0} (B);                 
            \draw[->] (B) to[bend left=20] node[midway, above]{c, +1} (C);   
            \draw[->] (C) to[bend left=20] node[midway, below]{b, 0}(B);   
        \end{tikzpicture}
    \end{center}
    State A cannot be reached from any other states in the MDP, thus, the decomposability condition is not satisfied. However, if there exists an abstraction \(\phi\), where \(\phi(A)=\phi(B)=Z\), then the abstract MDP becomes:
    
    \begin{center}
        \begin{tikzpicture}[
          >=Stealth,
          state/.style={circle, draw, minimum size=18pt, inner sep=0pt, font=\small}
        ]
            \node[state] (B) {B};
            \node[state, right=2.2cm of A] (Z) {Z};
            
            \draw[->] (B) to[bend left=20] node[midway, above]{c, +1} (Z);   
            \draw[->] (Z) to[bend left=20] node[midway, below]{b, 0}(B);   
        \end{tikzpicture}
    \end{center}

    Thus, the new abstract MDP satisfies the decomposability condition and there exists a disentangled reward function as list above.

\section{Training Details and Hyperparameters} \label{appendix:training}
    In this section, we show the comprehensive training details and hyperparameters. We use Soft Actor-Critic (SAC) as our Maximum Entropy Reinforcement Learning (MaxEnt RL) algorithm due to its efficient exploration, stability in continuous control tasks, and improved sample efficiency. By maximizing both cumulative reward and entropy, SAC promotes diverse and robust policies. For implementation, we use the SAC provided by the widely adopted Python library Stable-Baselines 3.

    To generate expert demonstrations, we first train SAC agents with 5 different random seeds in each source domain until convergence. The hyperparameters used for SAC training are listed in Table~\ref{table:tirl_sac_hyperparameter}. The unlisted hyperparameter remains the default setting in Stable-Baselines 3. After convergence, we collect 50 expert trajectories from each source domain. These expert trajectories are then used for training the transferable reward function via \tirl{} as well as other baseline methods.

    \begin{table}[!ht]
        \small          
        \centering
        
        \begin{tabular}{ccccc}
        \toprule
          &  Ant & HalfCheetah & FeedingSawyer & ScratchItchSawyer  \\
        \midrule
        Learning rate  & $3e^{-4}$ & $3e^{-4}$ & $3e^{-4}$ & $3e^{-4}$  \\
        Gamma & $0.99$ & $0.99$ & $0.99$ & $0.99$ \\
        Batch size & $1024$ & $1024$ & $1024$ & $1024$ \\
        Net arch & $[400, 300]$ & $[400, 300]$ & $[400, 400]$ & $[400, 400]$\\
        Buffer size & $1,000,000$ & $1,000,000$ & $1,000,000$ & $100,000$ \\
        Action noise & $\mathcal{N}(0, 0.2)$ & $\mathcal{N}(0, 0.2)$ & $\mathcal{N}(0, 0.2)$ & $\mathcal{N}(0, 0.25)$\\
        \bottomrule
        \end{tabular}
        \caption{Hyperparameter setting of SAC.}
        \label{table:tirl_sac_hyperparameter}
    \end{table}

    Next, we begin training \tirl{} and baselines. The hyperparameters used in each source domain are listed in Table~\ref{table:tirl_hyperparameter}-\ref{table:i2l_hyperparameter}. Notably, the SAC in \tirl{} adopts the same hyperparameters specified in Table~\ref{table:tirl_sac_hyperparameter}. The net arch represents the dimensions of the model for each layer, excluding the output layer. The reward function produces a single scalar value, which is activated by a Tanh function. The update step refers to the number of gradient updates performed during each iteration of the training process. All baselines are evaluated using their default hyperparameters. The only modification is for RIME, where the input to the discriminator is restricted to state-only features to ensure a fair comparison with our approach.
    
    \begin{table}[!ht]
    \small            
        \centering
        
        \begin{tabular}{cccc}
        \toprule
          & Ant & HalfCheetah & FeedingSawyer  \\
        \midrule
        \(\lambda_{\text{VAE}}\) & 1.0 & 1.0 & 1.0 \\
        \(\lambda_{\text{WGAN}}\) & 1.0 & 1.0 & 1.0 \\
        \(\lambda_{\mathcal{F}}\) & 1.0 & 1.0 & 1.0 \\
        \midrule
        \multicolumn{4}{c}{Reward Function Hyperparameter}\\
        \midrule
        Learning Rate & 3e-4 & 3e-4 & 5e-4 \\
        Batch Size & 2048 & 2048 & 2048 \\
        Weight Decay & 1e-3 & 1e-3 & 1e-3\\
        Net arch & [16, 16] & [16, 16] & [16, 16] \\
        Activation & Tanh & Tanh & Tanh \\
        Reward Update Steps & 10 & 10 & 10 \\
        \midrule
        \multicolumn{4}{c}{VAE and Discriminator Hyperparameter}\\
        \midrule
        Learning Rate & 3e-4 & 3e-4 & 5e-4 \\
        Batch Size & 2048 & 2048 & 2048 \\
        Weight Decay & 1e-3 & 1e-3 & 1e-3\\
        Encoder Net Arch & [32, 32, 32] & [32, 32, 32] & [16, 16] \\
        Encoder Activation & Tanh & Tanh & Tanh \\
        Abstraction Dimension & 16 & 10 & 4 \\
        Decoder Net Arch & [64, 64, 64] & [64, 64, 64] & [16, 16, 16] \\
        Decoder Activation & Tanh & Tanh & Tanh \\
        VAE Update Steps & 10 & 10 & 10 \\
        Discriminator Net Arch & [32, 32] & [32, 32] & [16, 16] \\
        Discriminator Activation & Tanh & Tanh & Tanh \\
        Disc Update Steps & 10 & 10 & 10 \\
        \bottomrule
        \end{tabular}     
        \caption{Hyperparameter setting of \tirl{} (Algorithm $\ref{algo:tirl-training}$).}
        \label{table:tirl_hyperparameter}
    \end{table}

    \begin{table}[!ht]
    \small            
        \centering
        
        \begin{tabular}{cccc}
        \toprule
          & Ant & HalfCheetah & FeedingSawyer  \\
        \midrule
        $g_\theta(s)$ network & [64, 64, 64] & [64, 64, 64] & [64, 64, 64] \\
        $h_{\phi^i}(s)$ network & [64, 64, 64] & [64, 64, 64] & [64, 64, 64] \\
        Learning rate & 3e-4 & 3e-4 & 3e-4 \\
        Batch size & 2048 & 2048 & 2048 \\
        Weight Decay & 1e-3 & 1e-3 & 1e-3 \\
        Activation & Tanh & Tanh & Tanh \\
        Discriminator gradient steps & 10 & 10 & 10 \\        
        \bottomrule
        \end{tabular}     
        \caption{Hyperparameter setting of AIRL-ME.}
        \label{table:airl-me_hyperparameter}
    \end{table}

    \begin{table}[!ht]
    \small            
        \centering
        
        \begin{tabular}{cccc}
        \toprule
          & Ant & HalfCheetah & FeedingSawyer  \\
        \midrule
        Policy network & [400, 300] & [400, 300] & [400, 300] \\
        Policy algorithm, lr, gradient-steps & PPO, 3e-4, 5 & PPO, 3e-4, 5 & PPO, 3e-4, 5 \\
        Discriminator network & [100, 100] & [100, 100] & [100, 100] \\
        Input to the discriminator & State-only & State-only & State-only \\
        Discriminator gradient-steps & 5 & 5 & 5 \\
        Gradient penalty term & 10 & 10 & 10 \\
        Batch size & 2048 & 2048 & 2048 \\
        Activation & Tanh & Tanh & Tanh \\
        \bottomrule
        \end{tabular}     
        \caption{Hyperparameter setting of RIME.}
        \label{table:rime_hyperparameter}
    \end{table}

    \begin{table}[!ht]
    \small            
        \centering
        
        \begin{tabular}{cccc}
        \toprule
          & Ant & HalfCheetah & FeedingSawyer  \\
        \midrule
        Wasserstein critic network & [64, 64, 64] & [64, 64, 64] & [64, 64, 64] \\
        Discriminator network & [64, 64, 64] & [64, 64, 64] & [64, 64, 64] \\
        Policy network & [400, 300] & [400, 300] & [400, 300] \\
        Wasserstein critic optimizer, lr, gradient-steps & Adam, 5e-5, 20 & Adam, 5e-5, 20 & Adam, 5e-5, 20 \\
        Discriminator optimizer, lr, gradient-steps & Adam, 3e-4, 5 & Adam, 3e-4, 5 & Adam, 3e-4, 5 \\
        Policy algorithm, lr & PPO, 1e-4 & PPO, 1e-4 & PPO, 1e-4 \\
        Batch size & 256 & 256 & 256 \\
        Activation & Tanh & Tanh & Tanh \\
        \bottomrule
        \end{tabular}     
        \caption{Hyperparameter setting of I2L.}
        \label{table:i2l_hyperparameter}
    \end{table}

    \begin{table}[!ht]
    \small            
        \centering
        
        \begin{tabular}{cc}
        \toprule
        Algorithm & URL  \\
        \midrule
        AIRL-ME & https://github.com/Ojig/Environment-Design-for-IRL \\
        RIME & https://github.com/JongseongChae/RIME\\
        I2L & https://github.com/tgangwani/RL-Indirect-imitation\\
        \bottomrule
        \end{tabular}     
        \caption{Source codes of baselines.}
        \label{table:baseline-link}
    \end{table}

    After training \tirl{}, we obtain a trained transferable reward function, which is then applied to the target task by replacing the original reward function. Consequently, when the agent interacts with the target task, it only has access to the trained reward function. We continue to use SAC with the hyperparameters specified in Table~\ref{table:tirl_sac_hyperparameter} for policy optimization in the target domain.
    
    \begin{figure*}[!ht]
             \centering
             \begin{subfigure}[b]{0.45\textwidth}
                 \centering
                 \includegraphics[width=\textwidth]{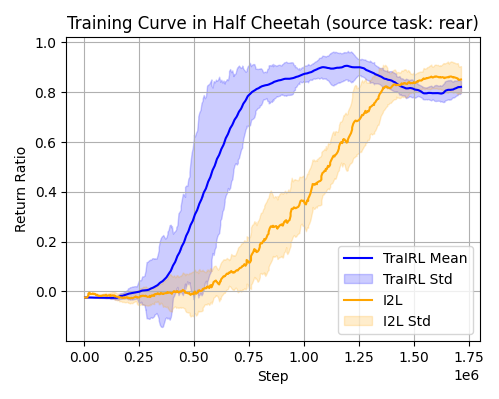}
                 \caption{\small Training curve in Half Cheetah (rear disabled). }
                 \label{fig:halfcheetah_source_1_training_curve}
             \end{subfigure}                 
             \begin{subfigure}[b]{0.45\textwidth}
                 \centering
                 \includegraphics[width=\textwidth]{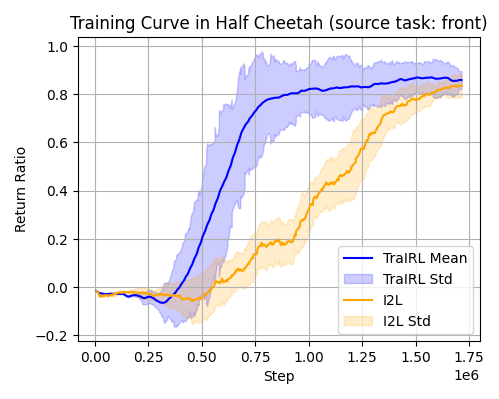}
                 \caption{Training curve in Half Cheetah (front disabled)}
                 \label{fig:halfcheetah_source_2_training_curve}
             \end{subfigure}                          
             \caption{\small Smoothed training curve for Half Cheetah in two source tasks. AIRL-ME and $f$-IRL perform poorly in the experiments and are therefore excluded from the comparison.}
             \label{fig:halfcheetah_training_curve}
    \end{figure*}

    \begin{figure*}[!ht]
            \centering
             \begin{subfigure}[b]{0.45\textwidth}
                 \centering
                 \includegraphics[width=\textwidth]{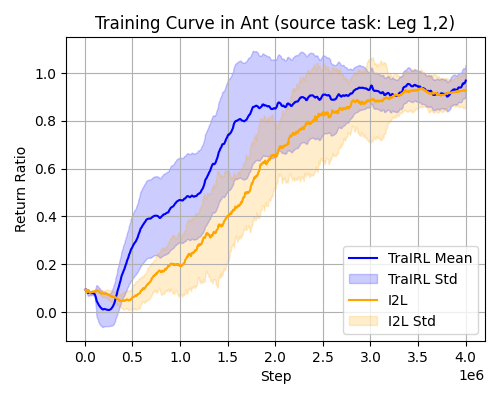}
                 \caption{\small Training curve in Ant (Leg 1 \& 2). }
                 \label{fig:ant1_training_curve}
             \end{subfigure}                 
             \begin{subfigure}[b]{0.45\textwidth}
                 \centering
                 \includegraphics[width=\textwidth]{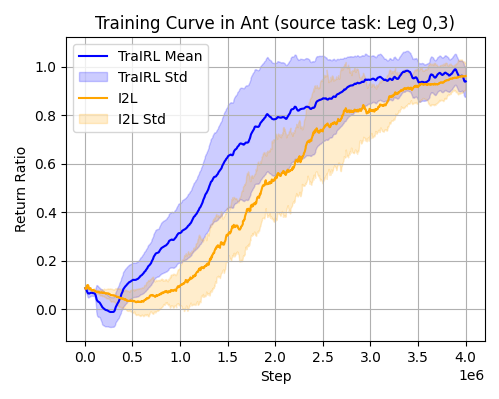}
                 \caption{Training curve in Ant (Leg 0 \& 3).}
                 \label{fig:ant2_training_curve}
             \end{subfigure}      
             \caption{\small Smoothed training curve for Ant in two source tasks. AIRL-ME and $f$-IRL perform poorly in the experiments and are therefore excluded from the comparison.}   
             \label{fig:ant_training_curve}
    \end{figure*}

    \begin{figure}[!ht]
        \centering
        \begin{subfigure}[b]{0.95\textwidth}
            \includegraphics[width=1.0\linewidth]{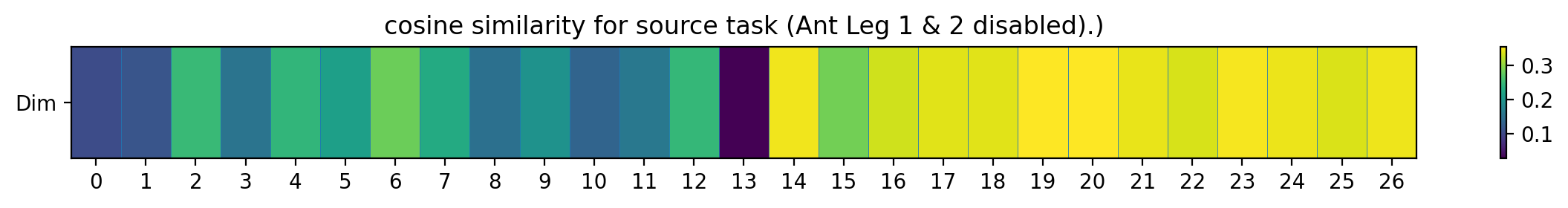}
        \end{subfigure}

        \begin{subfigure}[b]{0.95\textwidth}
            \includegraphics[width=1.0\linewidth]{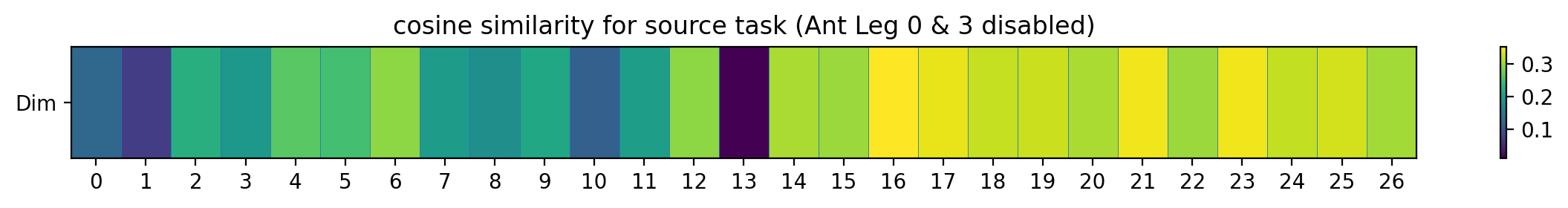}
        \end{subfigure}

        \begin{subfigure}[b]{0.95\textwidth}
            \includegraphics[width=1.0\linewidth]{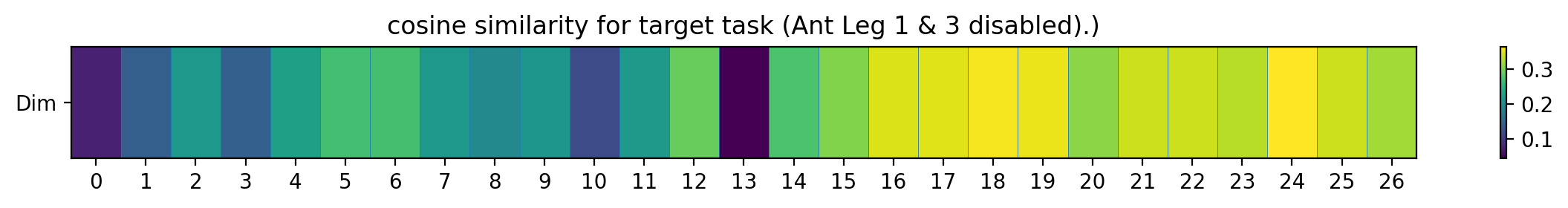}
        \end{subfigure}

        \begin{subfigure}[b]{0.95\textwidth}
            \includegraphics[width=1.0\linewidth]{pictures/cosine_similarity_for_target_task_Half_Cheetah.png}
        \end{subfigure}
        \caption{Visualization result for the abstraction sensitivity analysis via Cosine similarity.}
        \label{fig:semantic_analysis_cosine_similarity}
    \end{figure}
    
\newpage
\section{Extra Experiments and Clarifications} \label{appendix:extra experiments}    
    \subsection{Latent Sensitivity Analysis of Abstracted States} \label{appendix:semantic_analysis}
    In this section, we investigate the sensitivity of the learned abstract representations to individual dimensions of the ground state. This analysis identifies which state components are preserved by the encoder and which are treated as invariant or irrelevant information. The procedure for this sensitivity test is detailed in Algorithm \ref{algo:sensitivity_analysis}.

    \begin{algorithm} [!ht]
    \small
    \caption{Abstraction Sensitivity Analysis}
    \begin{algorithmic}[1]        
        \Require Learned encoder $p_{\phi}$, Trajectory buffer $\mathcal{B}$.
        \State Sample a batch of states $s$ from $\mathcal{B}$.
        \For{each dimension $d$ of the state space}
            \State Generate perturbed states $s^+ = s + \epsilon \cdot e_d$ and $s^- = s - \epsilon \cdot e_d$, where $\epsilon = 0.1$ and $e_d$ is the $d$-th basis vector.
            \State Compute latent embeddings: $z^+ = p_{\phi}(s^+), z^- = p_{\phi}(s^-)$.
            \State Calculate mean cosine similarity: $S_d = \mathbb{E}_s \left[ \frac{z^+ \cdot z^-}{\|z^+\| \|z^-\|} \right]$.
        \EndFor
        \State \Return Sensitivity profile $\{S_d\}$.
    \end{algorithmic}
    \label{algo:sensitivity_analysis}
    \end{algorithm}
    
    Figure \ref{fig:semantic_analysis_cosine_similarity} visualizes the absolute cosine similarity for each state dimension. In this framework, a \textbf{high similarity} score indicates that the encoder is \textbf{invariant} to perturbations in that dimension, suggesting the feature is filtered out as task-irrelevant. Conversely, a \textbf{low similarity} score indicates high \textbf{sensitivity}, revealing the features that the encoder prioritizes as essential components of the abstract representation.
    
    Our empirical results show that for the \textit{Ant} environment, the encoder exhibits the highest sensitivity (lowest similarity) at dimension 13. According to the Gymnasium documentation, this dimension corresponds to the \textbf{torso $x$-velocity} (forward movement). A similar pattern is observed in the \textit{HalfCheetah} environment, where the encoder is most sensitive to dimension 8, representing the \textbf{front tip velocity}. 
    
    These results demonstrate that the learned abstraction mechanism naturally converges on forward locomotion velocity as a primary task-relevant feature. This alignment confirms that the encoder successfully captures invariant properties across different task configurations, which is a critical requirement for robust generalization in reinforcement learning.

    \subsection{Integrate Abstract States into Other IRL Algorithms} \label{appendix:other_irl}
    In this section, we evaluate whether the abstraction representation \(\phi\) can be plugged into other IRL algorithms.
    
    \begin{table}[!ht]
        \centering

        \small
        \setlength\tabcolsep{6pt}
        \begin{tabular}{c|cc|cc}
            \toprule
            & \multicolumn{2}{c|}{\textbf{Sources}} & \multicolumn{2}{c}{\textbf{Targets}} \\
            & Source 1 & Source 2 & Target 1 & Target 2 \\
            \midrule
            AIRL with abstraction & 2579.61 $\pm$ 68.3 & \textbf{2966.92} $\pm$ \textbf{67.8} & 2271.65 $\pm$ 127.7 & 2766.01 $\pm$ 132.7 \\
            SFM with abstraction & \textbf{3002.88} $\pm$ \textbf{51.2} & 2851.09 $\pm$ 71.4 & 72.09 $\pm$ 18.3 & 52.78 $\pm$ 10.1 \\
            \tirl{} & 2714.18 $\pm$ 35.9 & 2936.52 $\pm$ 95.5 & \textbf{2917.92} $\pm$ 84.5 & \textbf{3156.54} $\pm$ \textbf{63.1} \\
            \midrule
            Expert & 3312.12 $\pm$ 304.3 & 3303.99 $\pm$ 341.0 & 3369.05 $\pm$ 216.8 & 3590.57 $\pm$ 158.2 \\
            \bottomrule
        \end{tabular}
        \caption{Comparison of \tirl{}, AIRL with abstraction, and SFM with abstraction as the feature function on source and target tasks.}
        \label{table:abstraction_ablation}
    \end{table}

    As shown in Table~\ref{table:abstraction_ablation}, both AIRL with abstraction and SFM with abstraction achieve higher returns than \tirl{} on the source tasks, demonstrating their ability to fit the training distributions more closely. However, these methods fail to preserve this advantage in the target tasks: SFM collapses almost entirely, and AIRL suffers a large drop in performance. In contrast, \tirl{} maintains strong generalization across targets. Compared to AIRL, \firl{} benefits from its separate reward function network, independent of the discriminator. The reward function is trained to maximize covariance with the discriminator’s output, effectively serving as a distillation process. This distillation enables the reward function to capture transferable information rather than overfitting to the source distributions, thereby improving generalization. For SFM with abstraction, the successor feature function is updated using the temporal consistency loss
    $$\mathbb{E}_{(s,a,s')\sim\mathcal{D}, \,a'\sim\pi_\mu(\cdot|s')}\Big[ \parallel\phi(s) + \psi_\theta(s', a') - \psi(s,a)\parallel^2_2  \Big],$$
    where \(\mathcal{D}\) is the buffer, \(\phi(s)\) is the base feature function, and \(\psi_\theta(s,a)\) is the successor feature function. Because this update depends directly on the current policy and state distribution, the learned successor features are tied to the specific dynamics and current policy. When the policy or state distribution shifts, as in the target tasks, the successor feature function becomes inaccurate. Since we do not retrain the successor features in the transfer setting, this mismatch results in poor generalization performance on unseen tasks.
    
    \subsection{Hyperparameter Sensitivity Analysis} \label{appendix:hyperparameter_sensitivity}
    In this section, we evaluate the sensitivity of \tirl{}'s performance on the coefficients $\lambda_{\text{GP}}$ and $\lambda_{\mathcal{D}}$. The coefficient $\lambda_{\text{GP}}$ controls the magnitude of the gradient penalty when updating the discriminator (Eq.~\ref{eqn:wgan_objective}), while $\lambda_{\mathcal{D}}$ regulates the strength of the regularization term when updating the encoder (Eq.~\ref{eqn:vae_objective}).
    
    ~\\    
    \textbf{Hyperparameter $\lambda_{\mathcal{D}}$}
    
    \begin{table}[!h]
        \label{ablation:lambda_D}
        \centering
        \small
        
        \begin{tabular}{c|ccc}
            \toprule
            & \multicolumn{2}{c}{\bf Sources} & \textbf{Target}\\
            & Run (rear disabled) & Run (front disabled) & Run (no disability) \\
            \midrule
            $\lambda_{\mathcal{D}} = 0.05$ & 4,323.23 $\pm$ 29.58 & 4,471.05 $\pm$ 56.21 & 4,541.09 $\pm$ 96.30 \\
            $\lambda_{\mathcal{D}} = 0.1$  & 4,404.07 $\pm$ 57.6 & 4,359.35 $\pm$ 99.2 & \textbf{5,835.11} $\pm$ \textbf{74.0} \\
            $\lambda_{\mathcal{D}} = 0.25$ & 4,430.26 $\pm$ 62.7 & 4,234.33 $\pm$ 84.0 & 4,548.23 $\pm$ 44.5 \\
            $\lambda_{\mathcal{D}} = 0.5$  & 3,806.92 $\pm$ 85.3 & 3,888.79 $\pm$ 52.5 & 4,088.95 $\pm$ 83.8 \\ 
            \midrule
            Expert  & 5,052.25 $\pm$ 25.4 & 5,499.07 $\pm$ 156.1 & 6,420.38 \(\pm\) 37.9 \\
            \bottomrule
        \end{tabular}  
        \caption{\small Mean cumulative reward with standard deviation in the \textbf{Half Cheetah} domain. $\lambda_{\mathcal{D}} = 0.1$ yields the highest cumulative reward in the target domain.}
        \label{table:ablation_lambda_D}
    \end{table}

    $\lambda_{\mathcal{D}}$ regulates the strength of the regularization term when updating the encoder (Eq.~\ref{eqn:vae_objective}), leading to a compact and generalizable abstracted state space. Table~\ref{table:ablation_lambda_D} reports the mean cumulative reward across different values of $\lambda_{\mathcal{D}}$ in the Half Cheetah domain. At $\lambda_{\mathcal{D}} = 0.1$, the model achieves the highest cumulative reward in the target environment (4745.59 $\pm$ 48.56), indicating that this value provides a balance for learning effective latent representations. Lowering $\lambda_{\mathcal{D}}$ to 0.05 slightly reduces performance in the target domain, suggesting that insufficient regularization may lead to suboptimal feature extraction. 
    Increasing $\lambda_{\mathcal{D}}$ beyond 0.1 results in a noticeable performance degradation. At $\lambda_{\mathcal{D}} = 0.25$, the reward declines to 4548.23 $\pm$ 44.49, and at $\lambda_{\mathcal{D}} = 0.5$, it further drops to 4088.95 $\pm$ 83.82. This decline suggests that excessive regularization constrains the encoder, limiting its ability to adapt to the target task. The performance reduction is also observed in source tasks, indicating that overly strong regularization affects overall learning stability.

\subsection{Insufficiency of Single Source Task} \label{appendix:single_source}
    In this section, we explore the insufficiency of a single-source task in training a transferable reward function. Recall that the objective of IRL is to match the state density or occupancy measure between the learner and expert policy. Therefore, if we only have a single source task, the learned reward function does not have to be generalized to any tasks except for the trained source task, even if the ground true reward function shares the same structure across tasks.  As shown in Table~\ref{table:single_source}, each reward function performs well on its corresponding training source task but generalizes poorly to others. This result highlights the necessity of using multiple diverse source tasks to capture a task-invariant reward structure for effective transfer learning.
    
    \begin{table}[!ht]
        \centering
        \small

        \begin{tabular}{cc|ccc}
            \toprule
            && \multicolumn{3}{c}{\bf Target}\\
            && Rear Disabled & Front Disabled & Normal \\
            \midrule
            \multirow{3}{*}{\bf Source} & Rear Disabled & \textbf{4,867.19} \(\pm\)\textbf{ 47.3} & 532.67 $\pm$ 312.2 & 3,158.79 $\pm$ 211.1 \\ 
                                        & Front Disabled& 831.98 $\pm$ 447.5 & \textbf{5,318.55} $\pm$ \textbf{83.9} & 3,351.9 $\pm$ 198.4 \\
                                        &     Normal    & 2,683.83 $\pm$ 412.7 & 3,017.66 $\pm$ 316.7 & \textbf{6,211.72} $\pm$ \textbf{42.0} \\
            \bottomrule
        \end{tabular}       
        \caption{Single source task transfer learning experiments on Half Cheetah.}
        \label{table:single_source}
    \end{table}

\subsection{Additional details about the 1-Wasserstein Distance Experiments}
    In this section, we detail the 1-Wasserstein distance experiments presented in Sec.~\ref{sec:benefit_of-abstraction}. The distance in the abstracted state space is computed using \tirl{}, while the corresponding distance in the ground state space is obtained from a variant of \firl{} in which the \(f\)-divergence is replaced by the 1-Wasserstein distance (see Appendix A.2 of \citet{firl}). In both cases, the 1-Wasserstein distance is estimated using the discriminator from a WGAN trained on expert trajectories from two source tasks. Although expert policies for target tasks are typically unavailable in standard IRL or transfer learning settings, we assume access to them in this experiment to enable the computation of the 1-Wasserstein distance defined in Theorem~\ref{theorem:trairl-applicability}, \(W_1(\rho_{\mathcal{T}^t}^*(z), \rho_{\mathcal{T}^i}^*(z))\), between the abstracted occupancy measures of the target and source tasks. The distances reported in Table~\ref{table:appendix_W_1} are computed from the state densities of the \textbf{expert policy}, either in the abstracted state space or the ground state space.
    
    \begin{table}[!ht]
        \small
        \centering  
        \begin{tabular}{ll|cc}
            \toprule
            && \multicolumn{2}{c}{\bf 1-Wasserstein Distance }\\
                                 & & Abstractions & Ground \\
            \midrule
            \multirow{3}{*}{\bf HalfCheetah} & Source (rear disabled) and Source (front disabled)  & {\bf 0.36} & 1.37 \\
            & Source (rear disabled) and Target (no disability)    &  {\bf 0.62} & 2.83 \\
            & Source (front disabled) and Target (no disability)    &  {\bf 0.55} & 2.10 \\
            \midrule
            \multirow{5}{*}{\bf Ant} & Source (Leg 1, 2 disabled) and Source (Leg 0, 3 disabled)    &  {\bf 0.33} & 1.78 \\
            & Source (Leg 1, 2 disabled) and Target (Leg 1, 3 disabled)    &  {\bf 0.71} & 2.82 \\
            & Source (Leg 1, 2 disabled) and Target (Leg 0, 2 disabled)    &  {\bf 0.79} & 2.97 \\
            & Source (Leg 0, 3 disabled) and Target (Leg 1, 3 disabled)    &  {\bf 0.81} & 3.00\\
            & Source (Leg 0, 3 disabled) and Target (Leg 0, 2 disabled)    &  {\bf 0.75} & 2.89\\
            \bottomrule
        \end{tabular} 
        \caption{Abstraction yields a smaller $W_1$ in the Half Cheetah and Ant, which is desirable.}
        \label{table:appendix_W_1}
    \end{table}

\subsection{Transfer Learning from Ant to Half Cheetah} \label{appendix:one-shot}
    \begin{figure}[!ht]
      \centering
       \includegraphics[width=0.95\textwidth]{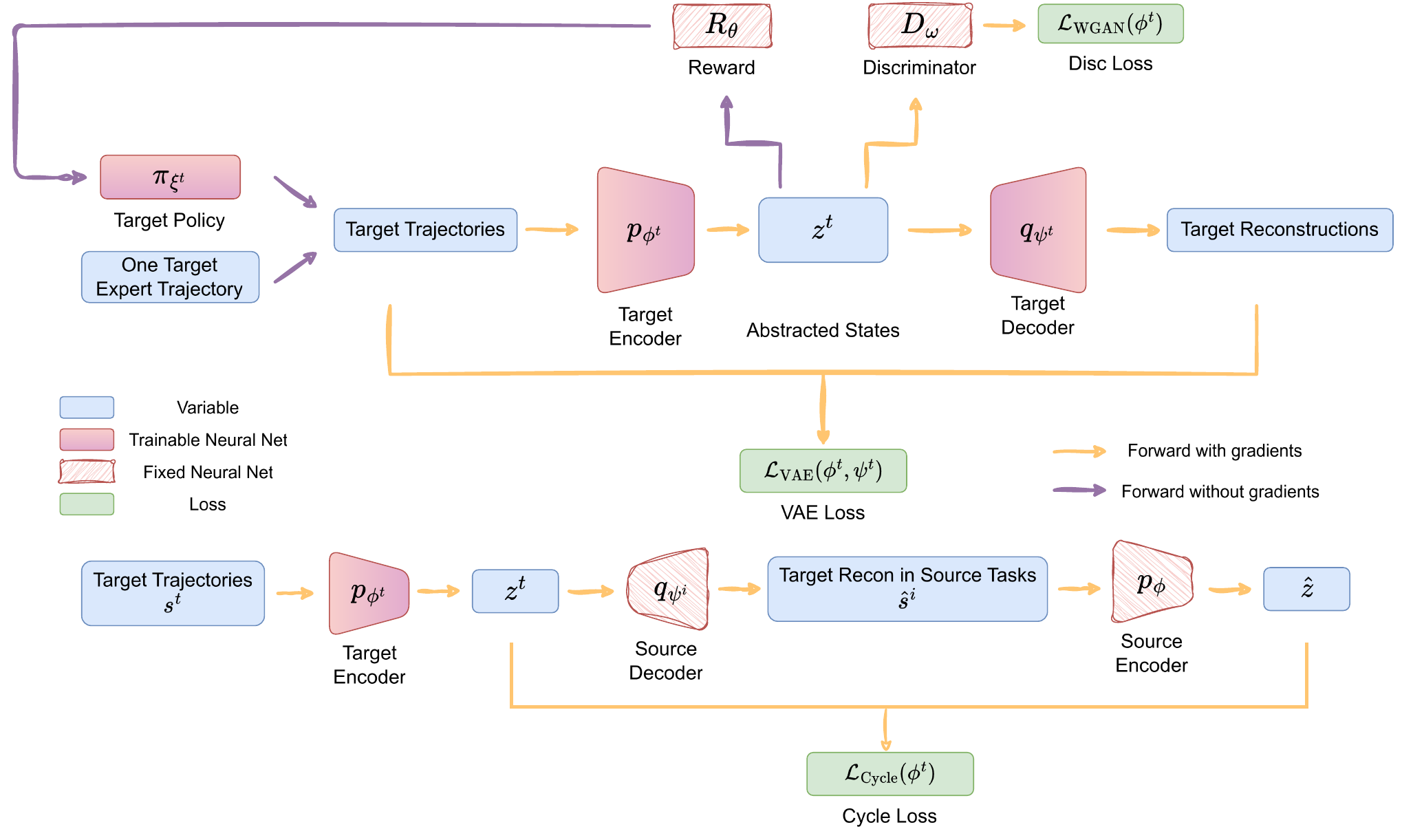}
       \caption{Overview of one-shot transfer learning in the target task.}
       \label{fig:cycle_diagram}
    \end{figure} 

    In this section, we describe the details for performing transfer learning from the source domain (Ant) to the target domain (Half Cheetah). The primary challenge in this cross-domain transfer arises from the mismatch between the ground state spaces of the two environments. Specifically, the Ant domain has a 27-dimensional state space, whereas the Half Cheetah domain has a 17-dimensional state space. Although both domains primarily consist of joint angles and velocities, the ranges of these values differ significantly between the two robots. As a result, directly applying the encoder trained in the source domain to the target domain leads to a substantial distribution shift, undermining the effectiveness of learned representations and transferred rewards. To mitigate the distribution shift, we introduce two methods, \textbf{one-shot} transfer learning and \textbf{zero-shot} transfer learning. Fig.~\ref{fig:cycle_diagram} shows the overview of transfer learning.
    
    \textbf{One-Shot Transfer Learning} ~~~We first introduce one-shot transfer learning.  Compared to training in the source domain, we have trained the reward function \(R_{\bm \theta}\), the discriminator \(D_\omega\), the encoder \(p_\phi\) and the decoder \(q_{\psi^i}\) for the source domain. Importantly, there is only \textbf{one} expert trajectory in the target domain, the so-called one-shot transfer learning. Specifically, we need to train an encoder \(p_{\phi^t}\), a decoder \(q_{\psi^t}\), and a policy \(\pi_{\xi^t}\) for the target domain. The VAE loss function remains the same as the training in the source domain, except for the single decoder in the target domain.
    
    \begin{align}
        \mathcal{L}_\text{VAE}(\phi^t, \psi^t) = \mathbb{E}_{z \sim p_{\bm \phi^t}(z^t|s^t)}\left[\log q_{\psi^t}(s^t|z)\right]   - \lambda_{\mathcal{D}}~\mathcal{D}_{\text{KL}}  \left(p_{\bm \phi^t}(z^t|s^t) \,\|\, p(z^t)\right).
    \end{align}

    Since the reward function \(R_\theta\) has been trained, we no longer need the reward loss. In terms of discriminator loss, the discriminator has also been trained. Therefore, the discriminator loss now only updates the encoder parameters:

    \begin{align} 
        \mathcal{L}_{\text{WGAN}}(\phi^t) = 
        & ~\mathbb{E}_{z \sim p_{\bm \phi^t}(z|s), s\sim\rho_L(s^t)} [D_{\bm \omega}(z)]  - \mathbb{E}_{z \sim p_{\bm \phi^t}(z|s), s\sim\rho_E(s^t)}[D_{\bm \omega}(z)] \nonumber \\
        & + \lambda_{\text{GP}}~\mathbb{E}_{z \sim \hat{\rho}(z^t)} \left[\left(\|\nabla_{z} D_{\bm \omega}(z)\|_2 - 1 \right)^2 \right]\Big),
    \end{align}

    Next, we introduce the novel cycle loss, illustrated in the overview in Fig.~\ref{fig:cycle_diagram}. The dimension of the abstracted state in the target domain is the same as that in the source domain. While the VAE loss and discriminator loss shape the abstract state space to be compact and optimality-aware, the cycle loss is designed to establish semantic alignment between the source and target domains. Specifically, it encourages consistency between the abstract state directly encoded from a target ground state and the abstract state obtained by first reconstructing that target abstracted state into the source domain and then encoding it. In doing so, the cycle loss ensures that abstract states from the target domain remain meaningfully aligned with those from the source, facilitating effective reward transfer despite differences in the ground state spaces.

    \begin{align}
        \mathcal{L}_\text{Cycle}(\phi^t) &= \mathcal{D}_\text{KL}\big(p_{\phi^t}(z^t|s^t) ~||~p_{\phi}(\hat{z}|\hat{s}^i)\big),
    \end{align}
    where \(\hat{s}^i\sim q_{\psi^i}(s^i|z^t)\).

    The overall loss function is the linear combination of the three loss functions:
    \begin{align}
        \label{eqn:one-shot-objective}
        \mathcal{L}(\phi^t, \psi^t) = \lambda_\text{VAE}\mathcal{L}_\text{VAE} +\lambda_\text{WGAN}\mathcal{L}_\text{WGAN} + \lambda_\text{Cycle}\mathcal{L}_\text{Cycle}
    \end{align}

    \textbf{Zero-Shot Transfer Learning} ~~~ Next, we introduce the zero-shot transfer learning setting, which requires the specification of goal criteria for expert behavior. To guide policy learning in the absence of expert demonstrations in the target domain, we apply a variant of the Hindsight Experience Replay (HER) technique to the learner trajectories. Unlike the original HER, which relabels achieved states as goals, our variant replaces the failed goal conditions in the learner trajectories with the expert-specified goal criteria. This relabeling produces more informative optimality signals, allowing the agent to learn effectively from suboptimal trajectories that would otherwise provide little to no feedback. For instance, in our zero-shot setting for the target domain (Half Cheetah), the goal criterion is achieving a high forward velocity. In the learner trajectories, where the agent typically exhibits low forward speed, we relabel the corresponding state features by replacing the observed speed with a high value that reflects expert-level performance. This substitution allows the learner to receive feedback aligned with the desired goal, thereby facilitating learning despite the absence of expert demonstrations in the target domain.

    The loss function of zero-shot transfer learning remains the same as Eq.~\ref{eqn:one-shot-objective}.

    \begin{table}[!ht]
        \centering
        \small

        \begin{tabular}{c|cc}
            \toprule
            & Half Cheetah (One-Shot) & Half Cheetah (Zero-Shot) \\
            \midrule
            Ant (Leg 1, 2) \&  (Leg 0, 3) (Source Tasks) &  5,378.78 \(\pm\) 61.7 & 4,821.79 $\pm$ 152.5 \\                 
            \bottomrule
        \end{tabular}      
        \caption{Transfer learning experiments on Half Cheetah.}
        \label{table:one-zero-shot}
    \end{table}

\subsection{Fine-Tuning with Reward Shaping} \label{appendix:reward_shaping}
    In Sec.~\ref{sec:assistive_gym}, we introduce that the robot must move its end effector precisely to a designated target area. However, the distributions of goal states differ between source and target tasks. Thus, reward shaping is  on top of the learned reward function \(R_{\bm\theta}\) in the target task by adding positive reward when goal states are reached:
    $\hat{R}(s) = R_{\bm\theta}(\phi(s)) + \mathbb{I}(s\in\mathcal{G})\cdot c$,
    where \(\mathbb{I}(s\in\mathcal{G})\) is the indicator function, \(\mathcal{G}\) is the set of goal states, and \(c\) is a positive constant. An experiment motivating this reward shaping is reported in Appendix \ref{appendix:reward_shaping}. Note that this reward shaping is applied consistently to \tirl{} as well as to all baseline methods to ensure a fair and comparable evaluation.Since the distributions of goal states differ between the source and target tasks, the learned reward alone is insufficient to guide the agent toward the new goal in the target task. Table~\ref{table:appendix_reward_shaping} illustrates the necessity of reward shaping by comparing performance with and without it. Note that reward shaping is applied consistently to all baseline algorithms as well as \tirl{}.

    \begin{table}[!ht]
        \centering
        \small

        \begin{tabular}{c|cc}
            & \multicolumn{2}{c}{Scratch Itch}\\           
            & without Reward Shaping & with Reward Shaping \\
            \midrule
            I2L     &  -19.55 \(\pm\) 3.90 & -10.07 $\pm$ 8.02 \\ 
            \tirl{} &  -20.13 \(\pm\) 5.11 & -3.82 $\pm$ 3.33 \\                 
        \end{tabular}          
        \caption{Reward shaping experiments on Assistive Gym environment. The reward function is learned in the source tasks (Feeding task 1 \& 2).}
        \label{table:appendix_reward_shaping}
    \end{table}

\subsection{Ablation Study}
    In this section, we conduct an ablation study on three objective functions, \(\mathcal{L}_\text{VAE}, \mathcal{L}_\text{WGAN}, \mathcal{L}_\mathcal{F}\), in the overall objective function, Eq.~\ref{eqn:trairl-objective}.

    \begin{table}[!ht]
        \centering
        
        \small
        
        \begin{tabular}{c|cc|c}
            \toprule
            & \multicolumn{2}{c}{\bf Sources} & \textbf{Target}\\
            & Run (rear disabled) & Run (front disabled) & Run (no disability) \\
            \midrule
            Without \(\mathcal{L}_\text{VAE}\)  & {\bf 4,099.25} $\pm$ {\bf 38.7} & {\bf 4,131.87} $\pm$ {\bf 47.8} & 1,083.63 $\pm$ 167.0 \\
            Without \(\mathcal{L}_\text{WGAN}\) & -133.89 $\pm$ 211.8 & -118.35 $\pm$ 259.4 & -181.2 $\pm$ 203.6 \\
            Without \(\mathcal{L}_\mathcal{F}\) & -148.1 $\pm$ 264.8 & -137.72 $\pm$ 213.7 & -189.0 $\pm$ 183.5 \\
            \midrule
            \tirl{}  & 4,404.07 $\pm$ 57.6 & 4,359.35 $\pm$ 99.2 & \textbf{5,835.11} $\pm$ \textbf{74.0} \\
            \bottomrule
        \end{tabular}
        \caption{Ablation Study}
        \label{table:ablation}
    \end{table}

    Table~\ref{table:ablation} presents an ablation study evaluating the contribution of the three key components in the \tirl{} objective: the VAE loss (\(\mathcal{L}_\text{VAE}\)), the discriminator loss (\(\mathcal{L}_\text{WGAN}\)), and the reward loss (\(\mathcal{L}_\mathcal{F}\)). Removing \(\mathcal{L}_\text{VAE}\) results in minimal impact on performance in the source tasks, but leads to a significant drop in the target task performance, suggesting that a shared abstracted state space is essential for transfer. In contrast, removing either \(\mathcal{L}_\text{WGAN}\) or \(\mathcal{L}_\mathcal{F}\) leads to a complete failure of learning across all tasks, with highly negative returns. This indicates the critical roles of adversarial training for enforcing optimality alignment and of reward fitting for capturing the expert policy structure. The full \tirl{} objective achieves strong performance across all tasks, particularly in the target domain, demonstrating that all three components are necessary for effective transfer learning.

    \subsection{Violation of the Structural Alignment Assumption in Theorem \ref{theorem:trairl-applicability}} \label{appendix:violation}
    As stated in Theorem~\ref{theorem:trairl-applicability}, a violation of the structural alignment assumption directly leads to a violation of Eq.~8. Intuitively, when this assumption is not satisfied, the abstract state density of the target task is shifted relative to that of the source tasks. Such a shift prevents the learned abstract reward from generalizing correctly, resulting in degraded performance on the target task. The shifted target abstract state density can be mitigated through one-shot transfer learning, as described in Appendix~\ref{appendix:one-shot}. We conducted an additional experiment to validate this claim, where source task 1: Ant (Leg 0 \& 1 disabled), source task 2: Ant (Leg 0 \& 3 disabled), and target task: Ant (Leg 1 \& 2 \& 3 disabled)

    \begin{table}[!ht]
        \centering
        \small

        \begin{tabular}{c|cc|c}
            \toprule
            & Source 1 & Source 2 & Target \\
            \midrule
            AIRL-ME& 2399.66 $\pm$ 89.1 & 2231.09 $\pm$ 87.6 & 169.00 $\pm$ 99.4 \\ 
            RIME   & 2471.90 $\pm$ 92.4 & 2308.14 $\pm$ 81.5   &   130.61 $\pm$ 108.3 \\
            I2L    & 2853.80 $\pm$ 73.8 & 2786.90 $\pm$ 79.4 & 149.52 $\pm$ 172.6 \\
            \tirl{} & 2815.93 $\pm$ 63.2 & 2936.52 $\pm$ 95.5 & 152.17 $\pm$ 153.1 \\
            \bottomrule
            Expert & 3391.57 $\pm$ 279.1 & 3303.99 $\pm$ 341.0 & 1655.42 $\pm$ 256.2 \\
            \bottomrule
        \end{tabular}
        \caption{Experiment on violation of the structural alignment assumption.}
        \label{table:failure_case}
    \end{table}

    In this setting, none of the source tasks contains optimality information about Leg 0, which becomes the only functional leg in the target task. This causes the abstract state distribution in the target to shift significantly from those in the source tasks, violating the structural alignment assumption. This result provides direct empirical support for Theorem \ref{theorem:trairl-applicability}: when the structural alignment assumption is violated, the abstract state density shifts, leading to poor generalization. After applying one-shot transfer learning, the issue of shifted abstract state density is mitigated, and the performance improves substantially.
    
    \subsection{Comprehensive Experiments} \label{appendix:comprehensive_experiments}
    In this section, we present additional experiments that incorporate a larger set of source and target tasks.

    \begin{table}[!ht]
        \centering

        \small
        \setlength\tabcolsep{4pt}
        \begin{tabular}{c|ccccc}
            \toprule
            & Leg 0,1 disabled & Leg 0,2 disabled & Leg 0,3 disabled & Leg 1,2 disabled & Leg 1,3 disabled \\
            \midrule
            AIRL\text{-}ME & 2,420.31 $\pm$ 75.2 & 2,361.22 $\pm$ 82.7 & 2,231.09 $\pm$ 67.6 & 2,389.65 $\pm$ 52.0 & 2,140.84 $\pm$ 75.3 \\
            RIME           & 2,691.04 $\pm$ 55.6 & 2,641.37 $\pm$ 91.2 & 2,708.14 $\pm$ 81.5 & \textbf{2,901.67} $\pm$ \textbf{49.9} & 2,590.71 $\pm$ 61.9 \\
            I2L            & 2,762.88 $\pm$ 64.1 & \textbf{2,801.55} $\pm$ \textbf{69.8} & 2,786.90 $\pm$ 79.4 & 2,831.28 $\pm$ 36.4 & 2,620.47 $\pm$ 84.5 \\
            \tirl{}         & \textbf{2,962.84} $\pm$ \textbf{49.5} & 2,774.88 $\pm$ 68.9 & \textbf{2,989.54} $\pm$ \textbf{53.0} & 2,844.94 $\pm$ 17.2 & \textbf{2,733.25} $\pm$ \textbf{56.9} \\
            \bottomrule
            Expert         & 3,127.77 $\pm$ 172.4 & 3,590.57 $\pm$ 158.2 & 3,303.99 $\pm$ 341.0 & 3,312.12 $\pm$ 304.3 & 3,369.05 $\pm$ 216.8 \\
            \bottomrule

        \end{tabular}
        \caption{Extended experiments on \textbf{source} tasks in the \textbf{Ant} domain (part 1).}
        \label{table:extended_source_ant_part1}
    \end{table}
    
    \begin{table}[!ht]
        \centering

        \small
        \setlength\tabcolsep{4pt}
        \begin{tabular}{c|ccccc}
            \toprule
            & Leg 2,3 disabled & Leg 0,1,2 disabled & Leg 0,1,3 disabled & Leg 0,2,3 disabled & Leg 1,2,3 disabled\\
            \toprule
            AIRL\text{-}ME & 2,302.44 $\pm$ 78.1 & 1,182.06 $\pm$ 50.3 & 1,160.92 $\pm$ 48.6 & 1,121.37 $\pm$ 46.9 & 1,131.58 $\pm$ 52.1 \\
            RIME           & 2,660.08 $\pm$ 79.6 & 1,301.77 $\pm$ 52.5 & \textbf{1,483.49 $\pm$ 36.8} & 1,241.62 $\pm$ 50.7 & 1,252.33 $\pm$ 44.2 \\
            I2L            & 2,721.63 $\pm$ 78.2 & 1,381.94 $\pm$ 40.1 & 1,362.15 $\pm$ 35.7 & \textbf{1,361.08} $\pm$ \textbf{38.9} & 1,332.76 $\pm$ 45.0 \\
            \tirl{}         & \textbf{2,802.07} $\pm$ \textbf{42.4} & \textbf{1,454.11} $\pm$ \textbf{29.6} & 1,418.54 $\pm$ 18.3 & 1,359.91 $\pm$ 23.1 & \textbf{1,361.90} $\pm$ \textbf{50.6} \\
            \bottomrule
            Expert         & 3,068.32 $\pm$ 209.7 & 1,716.93 $\pm$ 227.0 & 1,680.04 $\pm$ 199.8 & 1,652.81 $\pm$ 161.2 & 1,655.42 $\pm$ 256.2 \\

            \bottomrule
        \end{tabular}
        \caption{Extended experiments on \textbf{source} tasks in the \textbf{Ant} domain (part 2).}
        \label{table:extended_source_ant_part2}
    \end{table}
    
    \begin{table}[!ht]
        \centering

        \small
        \setlength\tabcolsep{6pt}
        \begin{tabular}{c|ccccc}
            \toprule
            & Leg 0 disabled & Leg 1 disabled & Leg 2 disabled & Leg 3 disabled & No leg disabled \\
            \midrule
            AIRL\text{-}ME & 2,412.67 $\pm$ 83.1 & 2,395.42 $\pm$ 102.7 & 2,368.19 $\pm$ 59.4 & 2,401.55 $\pm$ 97.2 & 2,680.14 $\pm$ 91.5 \\
            RIME           & 2,675.34 $\pm$ 72.8 & 2,659.48 $\pm$ 81.3 & 2,621.77 $\pm$ 69.5 & 2,670.91 $\pm$ 86.4 & 2,945.63 $\pm$ 68.2 \\
            I2L            & 2,752.18 $\pm$ 64.7 & 2,736.02 $\pm$ 65.0 & 2,701.36 $\pm$ 71.9 & 2,749.85 $\pm$ 79.6 & 3,028.77 $\pm$ 83.4 \\
            \tirl{}         & \textbf{3,241.03} $\pm$ \textbf{95.0} & \textbf{3,237.62} $\pm$ \textbf{64.3} & \textbf{3,217.80} $\pm$ \textbf{80.6} & \textbf{3,276.18} $\pm$ \textbf{75.7} & \textbf{3,546.96} $\pm$ \textbf{161.1} \\
            \bottomrule
            Expert         & 3,619.15 $\pm$ 139.7 & 3,550.75 $\pm$ 141.5 & 3,440.70 $\pm$ 123.7 & 3,584.58 $\pm$ 133.1 & 4,091.59 $\pm$ 159.2 \\

            \bottomrule
        \end{tabular}
        \caption{Extended experiments on \textbf{target} tasks in the \textbf{Ant} domain.}
        \label{table:extended_target_ant}
    \end{table}

    Tables~\ref{table:extended_source_ant_part1}–\ref{table:extended_target_ant} present an extended experiment in the Ant domain, which includes 10 source tasks and 5 target tasks. In the source tasks, I2L and RIME occasionally achieve slightly higher returns than \tirl{}, reflecting their ability to fit specific tasks. However, in the target tasks, \tirl{} clearly outperforms all baselines, demonstrating that the transferable abstract state space enables stronger generalization and more reliable reward transfer. This highlights the key advantage of \tirl{}: while other methods may match or surpass performance on the sources, only \tirl{} maintains superior performance when adapting to unseen target tasks.

    Another extended experiment is conducted in the Assistive Gym environment\footnote{https://github.com/Healthcare-Robotics/assistive-gym/wiki}, where source 1 (Feeding Sawyer with a static patient), source 2 (Feeding Sawyer with a tremor patient), source 3 (Feeding Sawyer with a static patient and a disabled wrist joint), source 4 (Feeding Sawyer with a tremor patient and a disabled wrist joint), source 5 (Scratch Itch Sawyer with a static patient), source 6 (Scratch Itch Sawyer with a tremor patient), source 7 (Scratch Itch Sawyer with a static patient and a disabled wrist joint), and source 8 (Scratch Itch Sawyer with a tremor patient and a disabled wrist joint); target 1 (Drinking Sawyer with a static patient), target 2 (Drinking Sawyer with a tremor patient), target 3 (Drinking Sawyer with a static patient with a static patient and a disabled elbow joint), and target 4 (Drinking Sawyer with a tremor patient and a disabled elbow joint).

    \begin{table}[!ht]
        \centering

        \small
        \setlength\tabcolsep{4pt}
        \begin{tabular}{c|ccccc}
            \toprule
            & Source 1 & Source 2 & Source 3 & Source 4 & Source 5 \\
            \midrule
            AIRL\text{-}ME & 3.8 $\pm$ 2.7 & 4.5 $\pm$ 6.9 & 1.1 $\pm$ 0.4 & 4.0 $\pm$ 2.3 & -5.6 $\pm$ 2.1 \\
            RIME           & 7.1 $\pm$ 7.1 & 7.8 $\pm$ 6.0 & 5.9 $\pm$ 3.3 & 7.2 $\pm$ 1.1 & -2.9 $\pm$ 4.8 \\
            I2L            & \textbf{10.9} $\pm$ \textbf{3.8} & 8.5 $\pm$ 3.1 & \textbf{10.0} $\pm$ \textbf{5.5} & 8.7 $\pm$ 6.2 & -2.5 $\pm$ 0.7 \\
            \tirl{}        & 9.3 $\pm$ 5.1 & \textbf{11.1} $\pm$ \textbf{4.2} & 8.7 $\pm$ 6.1 & \textbf{9.0} $\pm$ \textbf{6.1} & \textbf{-2.0} $\pm$ \textbf{1.6} \\
            \midrule
            Expert         & 11.2 $\pm$ 5.3 & 12.7 $\pm$ 4.2 & 9.54 $\pm$ 3.1 & 10.2 $\pm$ 4.0 & -1.1 $\pm$ 5.7 \\
            \bottomrule
        \end{tabular}
        \caption{Extended experiments on \textbf{source} tasks in the Assistive Gym (part 1).}
        \label{table:extended_source_assistive_gym_1}
    \end{table}
    
    \begin{table}[!ht]
        \centering

        \small
        \setlength\tabcolsep{4pt}
        \begin{tabular}{c|ccc}
            \toprule
            & Source 6 & Source 7 & Source 8 \\
            \midrule
            AIRL\text{-}ME & -6.9 $\pm$ 2.8 & -6.4 $\pm$ 1.9 & -6.7 $\pm$ 1.0 \\
            RIME           & -3.9 $\pm$ 3.7 & -3.4 $\pm$ 4.8 & -3.6 $\pm$ 5.9 \\
            I2L            & -3.1 $\pm$ 0.6 & \textbf{-2.9} $\pm$ \textbf{0.7} & -3.2 $\pm$ 1.6 \\
            \tirl{}        & \textbf{-2.7} $\pm$ \textbf{2.5} & -3.0 $\pm$ 2.6 & \textbf{-2.8} $\pm$ \textbf{2.5} \\
            \midrule
            Expert         & -2.3 $\pm$ 4.8 & -3.1 $\pm$ 4.9 & -2.9 $\pm$ 3.7 \\
            \bottomrule
        \end{tabular}
        \caption{Extended experiments on \textbf{source} tasks in the Assistive Gym (part 2).}
        \label{table:extended_source_assistive_gym_2}
    \end{table}
    
    \begin{table}[!ht]
        \centering

        \small
        \setlength\tabcolsep{4pt}
        \begin{tabular}{c|cccc}
            \toprule
            & Target 1 & Target 2 & Target 3 & Target 4 \\
            \midrule
            AIRL\text{-}ME & 12.4 $\pm$ 4.8 & 11.2 $\pm$ 2.5 & 10.9 $\pm$ 6.7 & 11.3 $\pm$ 5.6 \\
            RIME           & 18.9 $\pm$ 8.2 & 17.5 $\pm$ 10.0 & 16.8 $\pm$ 8.1 & 17.2 $\pm$ 7.2 \\
            I2L            & 20.7 $\pm$ 3.5 & 24.4 $\pm$ 5.2 & 22.8 $\pm$ 4.4 & 20.1 $\pm$ 3.3 \\
            \tirl{}        & \textbf{29.8} $\pm$ \textbf{6.1} & \textbf{27.2} $\pm$ \textbf{4.0} & \textbf{26.7} $\pm$ \textbf{6.2} & \textbf{25.0} $\pm$ \textbf{6.1} \\
            \midrule
            Expert         & 35.7 $\pm$ 11.7 & 30.5 $\pm$ 9.4 & 28.2 $\pm$ 11.0 & 25.7 $\pm$ 10.2 \\
            \bottomrule
        \end{tabular}
        \caption{Extended experiments on \textbf{target} tasks in the Assistive Gym.}
        \label{table:extended_target_assistive_gym}
    \end{table}

    Tables~\ref{table:extended_source_assistive_gym_1}–\ref{table:extended_target_assistive_gym} present extended experiments in the Assistive Gym domain, which include 8 source tasks and 4 target tasks. In the source tasks, I2L and RIME occasionally surpass \tirl{} on certain tasks, reflecting their strength in fitting task-specific structures. However, in the target tasks, \tirl{} consistently outperforms all baselines, achieving results that are much closer to expert performance. This demonstrates that the transferable abstract state space learned by \tirl{} enables stronger generalization and more reliable reward transfer. The results highlight the key advantage of \tirl{}: while baselines can sometimes match or exceed performance in the sources, only \tirl{} sustains superior generalization when transferring to unseen target tasks.

    As the number of source tasks increases, the reward learned by \tirl{} becomes more transferable because the abstraction is trained on a richer and more diverse set of trajectories. This diversity allows the abstract state space to capture higher-level features that are invariant across a broader range of variations, reducing the chance of overfitting to any single task. Consequently, the learned reward generalizes more effectively to unseen targets, as it encodes task-independent structure rather than idiosyncratic details. In practice, incorporating more source tasks also improves robustness, since the abstraction must reconcile multiple dynamics and objectives, leading to a more stable and transferable reward representation.

\section{Computing Resources}
All experiments were conducted on a desktop machine running Ubuntu 20.04, equipped with an Intel Core i7-10700K CPU, 32 GB of RAM, an NVIDIA RTX 3070 GPU, and CUDA 12.6.

\section{Visualization of Abstracted State Space} \label{appendix:tsne}
    \begin{figure*}[!ht]
         \centering            
         \begin{subfigure}[b]{0.45\textwidth}
             \centering
             \includegraphics[width=\textwidth]{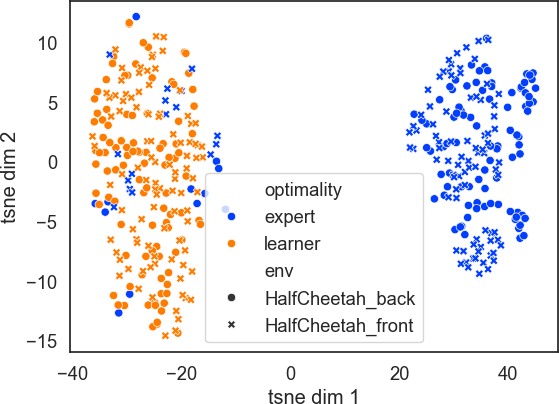}
             \caption{\small Abstraction}
             \label{fig:appendix_halfcheetah_visual_abstraction}
         \end{subfigure}
         \begin{subfigure}[b]{0.45\textwidth}
             \centering
             \includegraphics[width=\textwidth]{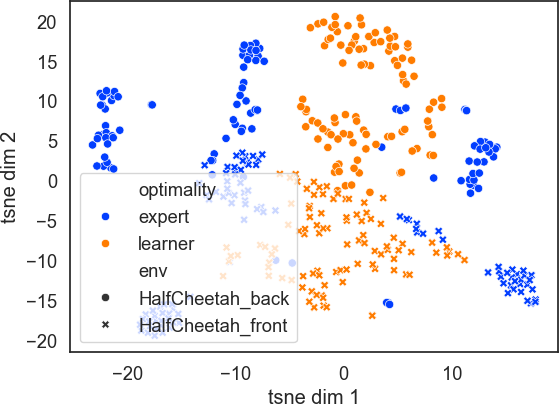}
             \caption{\small Ground}
             \label{fig:appendix_halfcheetah_visual_state}
         \end{subfigure}
         \caption{Visualization of \textbf{Half Cheetah} domain by t-SNE.}
         \label{fig:appendix_halfcheetah_visua}
    \end{figure*}
    
    \begin{figure*}[!ht]
        \centering
        \begin{subfigure}[b]{0.45\textwidth}
         \centering
         \includegraphics[width=\textwidth]{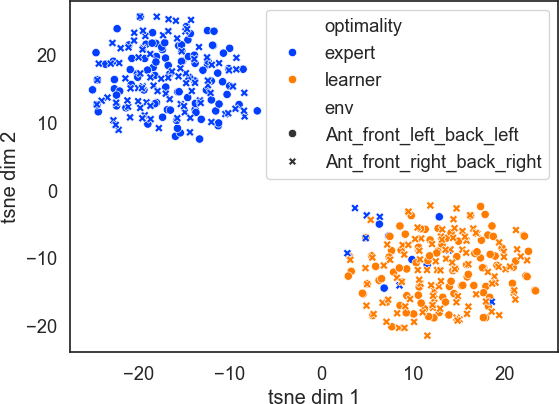}
         \caption{\small Abstraction}
         \label{fig:appendix_ant_visual_abstraction}
        \end{subfigure}
        \begin{subfigure}[b]{0.45\textwidth}
         \centering
         \includegraphics[width=\textwidth]{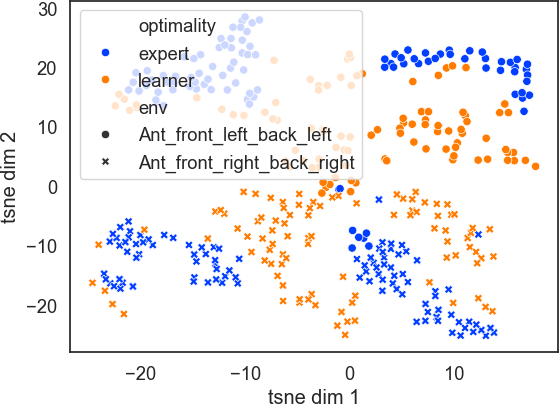}
         \caption{\small Ground}
         \label{fig:appendix_ant_visual_state}
        \end{subfigure}
        \caption{Visualization of \textbf{Ant} domain by t-SNE.}
        \label{fig:appendix_ant_visua}
    \end{figure*}

    Fig.~\ref{fig:appendix_halfcheetah_visua} presents t-SNE visualizations comparing the distributions of expert and learner states across two Half Cheetah tasks in both the ground and abstracted state spaces. In Fig.~\ref{fig:appendix_halfcheetah_visual_abstraction}, the expert and learner trajectories form well-separated clusters, orange for the learner and blue for the expert, while the trajectories from the two source tasks, Half Cheetah with rear legs disabled (circle) and with front legs disabled (cross), appear largely overlapping in the abstract state space. The overlapping of trajectories across different tasks and the clear separation between the expert and the learner demonstrate that the abstract state space captures task-invariant features while preserving optimality information useful for reward learning. In Fig.~\ref{fig:appendix_halfcheetah_visual_state}, both the expert and learner trajectories, as well as the two source tasks, form clearly separated clusters. This indicates the presence of task-specific features in the ground state space, which hinders the generalization of learned rewards from source tasks to the target task. The same conclusion can be made in the Ant domain in Fig.~\ref{fig:appendix_ant_visua}.

    \section{Limitations} \label{appendix:limitations}
    While \tirl{} demonstrates strong generalization across related tasks, several limitations remain. First, the method assumes that source and target tasks share sufficient structural similarity. When this assumption is violated, such as in cases with large differences in state distributions or dynamics, the transferability of the learned reward may degrade. Balancing the training loss between source tasks also poses challenges, especially when their difficulty or distribution differs significantly. An imbalance can cause the abstract representation to overfit to one source task, reducing its effectiveness in the target.

    Second, although \tirl{} enables zero-shot transfer within the same task family, transferring to structurally different domains (e.g., from Ant to Half Cheetah) requires additional adaptation. This often involves learning mappings between ground states or introducing domain-specific constraints to mitigate distribution shifts.
    
    Third, learning a reward function based on a common abstract representation may in some cases hurt performance. If the abstraction suppresses task-specific features that are critical for solving the target task, the resulting reward may fail to induce effective policies. As we showed in Sec \ref{sec:assistive_gym}, a goal-specific reward should be added on top of the abstract reward.
    
    Finally, \tirl{} introduces multiple interacting components—encoder, discriminator, and reward function—each with their own hyperparameters. Tuning these across tasks can be non-trivial. Overly complex discriminators or reward models may overfit to source tasks, harming generalization. Moreover, performance is sensitive to the balance among its three objectives: abstraction quality (VAE), optimality separation (discriminator), and reward alignment (reward function). Achieving this balance remains a key challenge for stable and effective training.

\end{document}